\documentclass[nohyperref]{article}

\usepackage{microtype}
\usepackage{graphicx}
\usepackage{subfigure}
\usepackage{booktabs} 

\usepackage{hyperref}
\usepackage{multirow}

\usepackage[accepted]{icml2022}

\usepackage{amsmath}
\usepackage{amssymb}
\usepackage{mathtools}
\usepackage{amsthm}

\usepackage[capitalize,noabbrev]{cleveref}

\theoremstyle{plain}
\newtheorem{theorem}{Theorem}[section]

\theoremstyle{definition}

\newtheorem{assumption}[theorem]{Assumption}
\theoremstyle{remark}

\usepackage[textsize=tiny]{todonotes}

\icmltitlerunning{DAPDAG}

\begin{document}

\twocolumn[
\icmltitle{DAPDAG: Domain Adaptation via \\
Perturbed DAG Reconstruction}



\icmlsetsymbol{equal}{*}

\begin{icmlauthorlist}
\icmlauthor{Yanke Li}{eth}
\icmlauthor{Hatt Tobias}{eth}
\icmlauthor{Ioana Bica}{ox}
\icmlauthor{Mihaela van der Schaar}{cam}
\end{icmlauthorlist}

\icmlaffiliation{eth}{ETH Zurich}
\icmlaffiliation{cam}{University of Cambridge}
\icmlaffiliation{ox}{University of Oxford}

\icmlcorrespondingauthor{Yanke Li}{yankli@student.ethz.ch}


\icmlkeywords{Machine Learning, ICML}

\vskip 0.3in
]



\printAffiliationsAndNotice{\icmlEqualContribution} 

\begin{abstract}
Leveraging labelled data from multiple domains to enable prediction in another domain without labels is an important, yet challenging problem. To address this problem, we introduce the framework DAPDAG (\textbf{D}omain \textbf{A}daptation via \textbf{P}erturbed \textbf{DAG} Reconstruction) and propose to learn an auto-encoder that undertakes inference on population statistics given features and reconstructing a directed acyclic graph (DAG) as an auxiliary task. The underlying DAG structure is assumed invariant among observed variables whose conditional distributions are allowed to vary across domains led by a latent environmental variable $E$. The encoder is designed to serve as an inference device on $E$ while the decoder reconstructs each observed variable conditioned on its graphical parents in the DAG and the inferred $E$. We train the encoder and decoder jointly through an end-to-end manner and conduct experiments on both synthetic and real datasets with mixed types of variables. Empirical results demonstrate that reconstructing the DAG benefits the approximate inference and furthermore, our approach can achieve competitive performance against other benchmarks in prediction tasks, with better adaptation ability especially in the target domain significantly different from the source domains.
\end{abstract}

\section{Introduction} 
Domain adaptation (DA) concerns itself with a scenario where one wants to transfer a model learned from one or more labelled source datasets, to a target dataset (which can be labelled or unlabelled) drawn from a different but somehow related distribution. In many settings, a wealth of data may exist, contained in several datasets collected from different sources, such as different hospitals, yet the target domain has few labels available due to possible lag or in-feasibility on data collection. Knowing what information can be transferred across domains and what needs to be adapted becomes the key to leveraging these datasets when presented unlabelled dataset from another domain, which has attracted significant attention in the machine learning community. In this paper, we revisit the problem of unsupervised domain adaptation (UDA), where the target dataset is unlabelled, under the same feature space with multiple source domains. To avoid over repetition, the word ``environment'' and ``domain'' are used interchangeably in the following paragraphs. 

For UDA, there have been various approaches developed and most of them can be summarised into two main categories - either learning an invariant representation of features with implicit alignment over source domains and the target domain, or utilising the underlying causal assumptions and knowledge that provide clues on the source of distribution shift for better adaptation. Despite the success of invariant representation methods in visual UDA tasks \citep{wang2018deep,deng2019cluster,kang2019contrastive,lee2019drop,liu2019transferable,jiang2020implicit}, its black-box nature remains vague locally and causes issue in some situations \citep{zhao2019learning}. Exploring the underlying causal structure and properties may help add more interpretability and make predictions across different domains more robust \footnote{Robustness refers to generalisation ability of model to unseen data}. In 
most settings, the causal structure of variables (both the features $\mathbf{X}$ and the label $Y$) are assumed to remain constant across domains and the label has a fixed conditional distribution given causal features \citep{scholkopf2012causal,magliacane2017domain}. In this work, we expect to capture similar invariant structural information but with conditional shift. More specifically, we cast the data generating process of distinct domains as a probability distribution with a continuous latent variable $E$ that perturbs the conditional distributions of observed variables. An auto-encoder approach is proposed to capture this latent $E$, utilising structural regularisation to facilitate sparsity and acyclicity among variable relationships. Our model is expected to be able to make inference on $E$, which is further used to adjust for domain shift in prediction. To accomplish this, the encoder structure is designed to approximate the posterior of $E$, drawing insights from methods of deep sets \citep{zaheer2017deep} and Bayesian inference \citep{maeda2020meta}, while the decoder aims to reconstruct all observed variables in a DAG taking the inferred $E$. 

\paragraph{Contributions}
The main contributions of this paper are three-fold:
\begin{itemize}
	\item We present a framework consisting of a encoder for the approximate inference on domain-specific variable $E$, and a decoder to reconstruct mixed-type data including continuous and binary variables. A novel training strategy is proposed to train our model: weighted stochastic domain selection, which enables inter and intra-domain validation during training. 
	\item We provide a generalisation bound on the decoder in our structure with mixed-type data, validating the training loss form to some extent.
    \item We validate our method with experiments on both simulated and real-world datasets, demonstrating the effectiveness of DAG reconstruction and performance gain of our approach in prediction tasks against benchmarks.
\end{itemize}

\paragraph{Related Work}  Since we are more interested in causal methods for UDA, review on other UDA methods would not be discussed in this paper. For more detailed reviews on general DA methods, please refer to \cite{quinonero2009dataset} and \cite{pan2009survey}. Various approaches have been proposed in causal UDA yet most of them can be categorised into three classes: (1) Correcting distribution shift by different scenarios of UDA according to underlying causal relations between $\mathbf{X}$ and $Y$ (e.g. to estimate the target conditional distribution as a linear mixture of source domain conditionals by matching the target-domain feature distribution) \citep{scholkopf2012causal,zhang2015multi,stojanov2019data};
(2) Identifying invariant subset of variables across domains for robust prediction \citep{magliacane2017domain,rojas2018invariant}; (3) Augmenting causal graph by considering interventions or environmental changes as exogenous (context) variables which affect endogenous (system) variables and implementing joint causal inference (JCI) on these augmenting graphs \citep{mooij2020joint,zhang2020domain}. Our approach is closest to the third class, by introducing an latent perturbation variable $E$ that induces conditional shift of observed variables. The resulting graph may not be causal any more, nevertheless our focus is the DAG representation of the whole distribution, which enforces sparsity and acyclicity for better learning of $E$. 

Our entire framework also takes resemblance with meta learning (for a survey on this, please see \cite{vilalta2002perspective,vanschoren2018meta}). In our setting, the objective is to learn an algorithm from different training tasks (domains) and to apply the algorithm to a new  task (domain). \cite{maeda2020meta} introduces an auto-encoder model to learn the latent embedding of different tasks under the Bayesian inference framework, which has similar mechanism with ours except that our decoder aims to reconstruct all variables in a DAG instead of only the target variable. There also exist a few works using meta-learning approach to handle variant causal structures across domains \citep{nair2019causal,dasgupta2019causal,ke2020amortized,lowe2020amortized}. Since our approach assumes an invariant DAG structure, we would not dive deeper into those methods although they may provide inspiring reference for our future work.  

\section{Preliminaries}
Learning a casual DAG is a hard problem that needs exhaustive search over a super-exponential combinatorial DAG space, which becomes impossible to deal with in high-dimensional case. However, recent advances in structure learning \citep{zheng2018dags,yu2019dag,zhang2019d,lachapelle2019gradient,yang2020causalvae,zheng2020learning} reduce the original combinatorial optimisation problem to a continuous optimisation by using a novel acyclicity constraint, which accelerate the learning and provide more inspirations. Some works have been extended to more complex settings including structural learning across non-stationary environments \citep{ghassami2018multi,bengio2019meta,ke2019learning}. Despite difference in implementation, above methods use end-to-end optimisation with standard gradient-descent methods that are on-the-shelf. In our work, we take the advantage of continuous optimisation methods and emphasise on NO-TEARS methods \citep{zheng2018dags,zheng2020learning} that can be better integrated into the deep learning framework. We consider learning a DAG as a auxiliary task to improve model's generalisation and robustness \citep{kyono2020castle}, contributing to the better learning of latent variable $E$ in the meantime.  

An example is introduced below to recap the basic idea of the NO-TEARS method. Suppose we want to learn a linear SEM (Structural Equation Model) with the form $\mathbf{X} = \mathbf{X}\mathbf{B} +\mathbf{\epsilon}$ where $\mathbf{\epsilon}$ is the random noise variable and $\mathbf{B} \in \mathbb{R}^{d\times d}$ is the weighted adjacency matrix. Then it can be proved that:
\begin{equation}
\mathbf{B} \text{ is a DAG} \Leftrightarrow h(\mathbf{B}) = Tr(e^{\mathbf{B}\odot \mathbf{B}})-d = 0    
\end{equation}
where $\odot$ is the Hadamard product $[\mathbf{B}\odot \mathbf{B}]_{ij}=\mathbf{B}_{ij}^2$.

For formal proof, please refer to \cite{zheng2018dags}. This formulation converts learning a linear DAG into a non-convex optimisation problem: 
\begin{align}
\min\limits_{\mathbf{B}} \quad \mathcal{L}(\mathbf{B})= \frac{1}{2n}||\mathbf{X}-\mathbf{X}\mathbf{B}||_F^2 +\lambda |vec(\mathbf{B})|_1 \nonumber \\
\quad\text{subject to} \quad h(\mathbf{B}) = 0
\end{align}
In \cite{zheng2018dags}, they solve the above problem by augmenting quadratic penalty and using Lagrangian method:
\begin{equation}
\min\limits_{\mathbf{B}} \quad \mathcal{L}(\mathbf{B}) + \frac{\rho}{2}|h(\mathbf{B})|^2 + \alpha h(\mathbf{B})    
\end{equation}
where $\rho$ is the penalty coefficient and $\alpha$ is the Lagrangian Multiplier. A further extension of this conversion has proposed by \cite{zheng2020learning} to the case of general non-parametric DAGs. Please refer appendix \ref{dag-notears} for a detailed illustration.

\section{Methodology}
\subsection{Formulation}

\paragraph{Problem Setting}
Let $Y$ be the target variable and $\mathbf{X} \in \mathbb{R}^d$ be features. We consider $M$ labeled datasets from different source domains, i.e. $(\mathbf{X}_i^m,Y_i^m)_{i=1}^{n_m} \sim \mathbb{P}^m$ where $m\in \{1,2,...,M\}$ represents the domain index, $\mathbb{P}^m$ stands for the probability distribution of $(\mathbf{X},Y)$ in domain $m$ and $n_m$ is the dataset size of domain $m$. Our objective is to predict $(Y_i^{\tau})_{i=1}^{n_{\tau}}$ given $(\mathbf{X}_i^{\tau})_{i=1}^{n_{\tau}}$ from the target domain $\tau$ without labels.   

\paragraph{Basic Assumptions} Let $\tilde{\mathbf{X}} = (\mathbf{X},Y) \in \mathbb{R}^{d+1}$ be observed variables, we assume:
\begin{itemize}
	\item Besides $\tilde{\mathbf{X}}$, there is a latent environmental variable $E$ controlling the distribution shift of observed variables. For each domain, $E$ is sampled from its prior $\mathcal{N}(0,\sigma_e^2)$ and fixed for data generation.
	\item Observed data are generated according to a perturbed DAG: the conditional distribution of $\tilde{X}_j$ given its parents and $E$ follows an exponential family distribution in the form of:
	\begin{align}
	p(\tilde{X}_j|\tilde{X}_{Pa(j)},E) &= \exp(\eta (\tilde{X}_{Pa(j)},E)\cdot T(\tilde{X}_j) \nonumber \\
	&+ A(\tilde{X}_{Pa(j)},E)+B(\tilde{X}_j))    
	\end{align}
	    
	where $\eta(\cdot)$, $A(\cdot)$ and $B(\cdot)$ are functions.
\end{itemize}

\begin{figure}[h]
	\centering
	\includegraphics[scale=0.6]{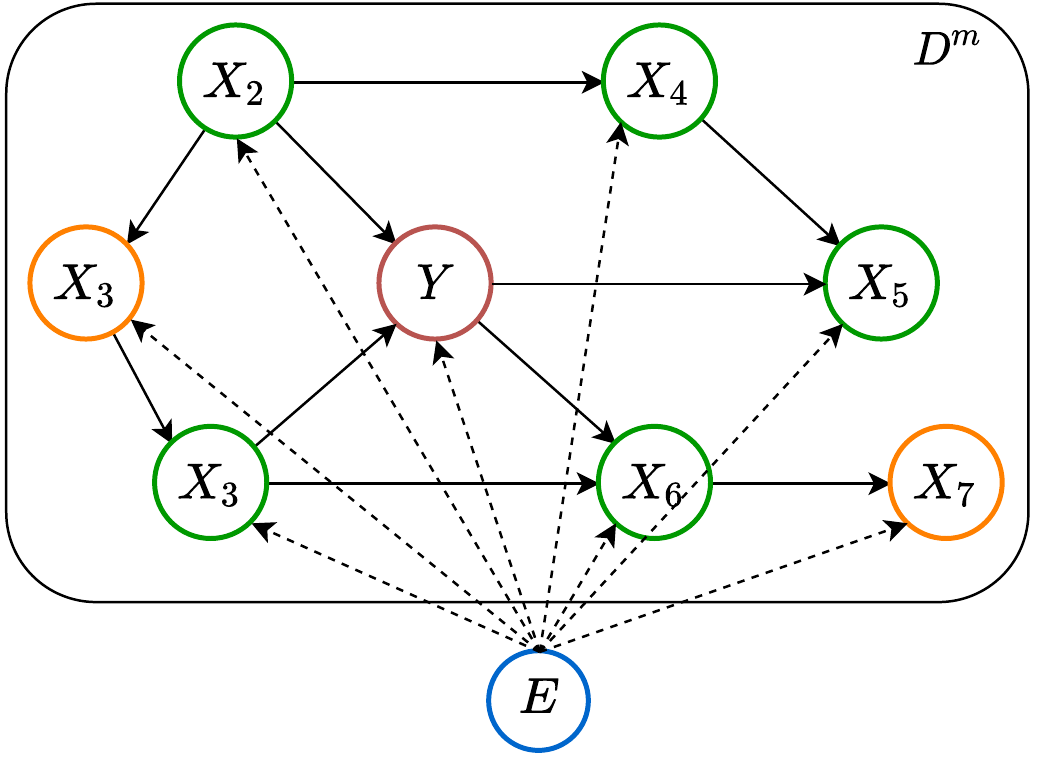}
	\caption[Perturbed DAG across Different Domains]{Perturbed DAG across Different Domains: for each domain, an environmental variable $E$ is generated and fixed for that domain, then all observed variables are sampled according to the DAG and $E$.}
	\label{fig:aug}
\end{figure}

\paragraph{Perturbed DAG} We assume a perturbed DAG where a joint environmental variable $E$ will influence the conditional distribution of an observed variable across domains. The illustration of this perturbed DAG is shown in Figure \ref{fig:aug}.

\subsection{Model}
\begin{figure*}[ht!]
	\centering
	\includegraphics[scale=0.48]{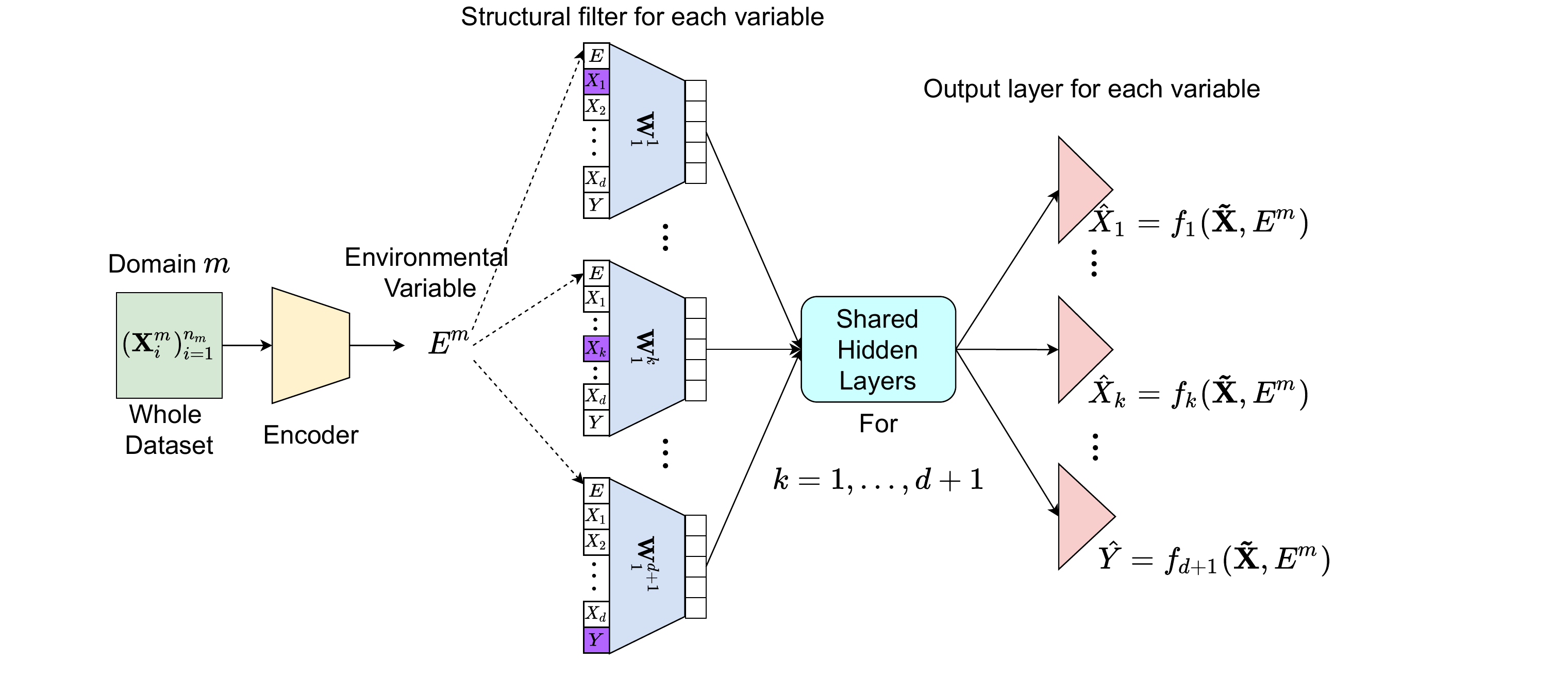}
	\caption[Model Structure]{Overview of model structure: $\mathbf{W}_1^k$ is the structural filter of the $k$-th variable comprised of a $(d+1)\times h$ matrix where $h$ is the number of hidden units in the hidden layer. To reconstruct the $k$-th variable, all entries in the $k$-th row of $\mathbf{W}_1^k$ are set to 0 to avoid using itself for reconstruction.}
	\label{DAPDAG}
\end{figure*}

We expect a model that is able to well capture the difference between $E$ of different domains, and then adapt to the change accordingly. So how to properly encode an empirical distribution to a statistics becomes the cornerstone of our model. Considering similarity with the goal of classical statistical estimation methods such as Maximum Likelihood Estimation (MLE), our objective is to learn an estimation device that can output the estimated $E$ for each domain taking its samples as input. The model has an auto-encoder  architecture, with an encoder to take the whole domain sample to approximate $E$ and a decoder to reconstruct each feature according its graphical parents and $E$. The latter bears resemblance with CASTLE \citep{kyono2020castle} except that $E$ is used for reconstruction. Figure~\ref{DAPDAG} sketches the general model architecture which consists of a domain encoder, a set of structural filters, shared hidden layers and separate output layers. We now explain each part in detail.

\subsubsection{Domain Encoder} 
\begin{figure}[h]
	\centering
	\includegraphics[scale=0.44]{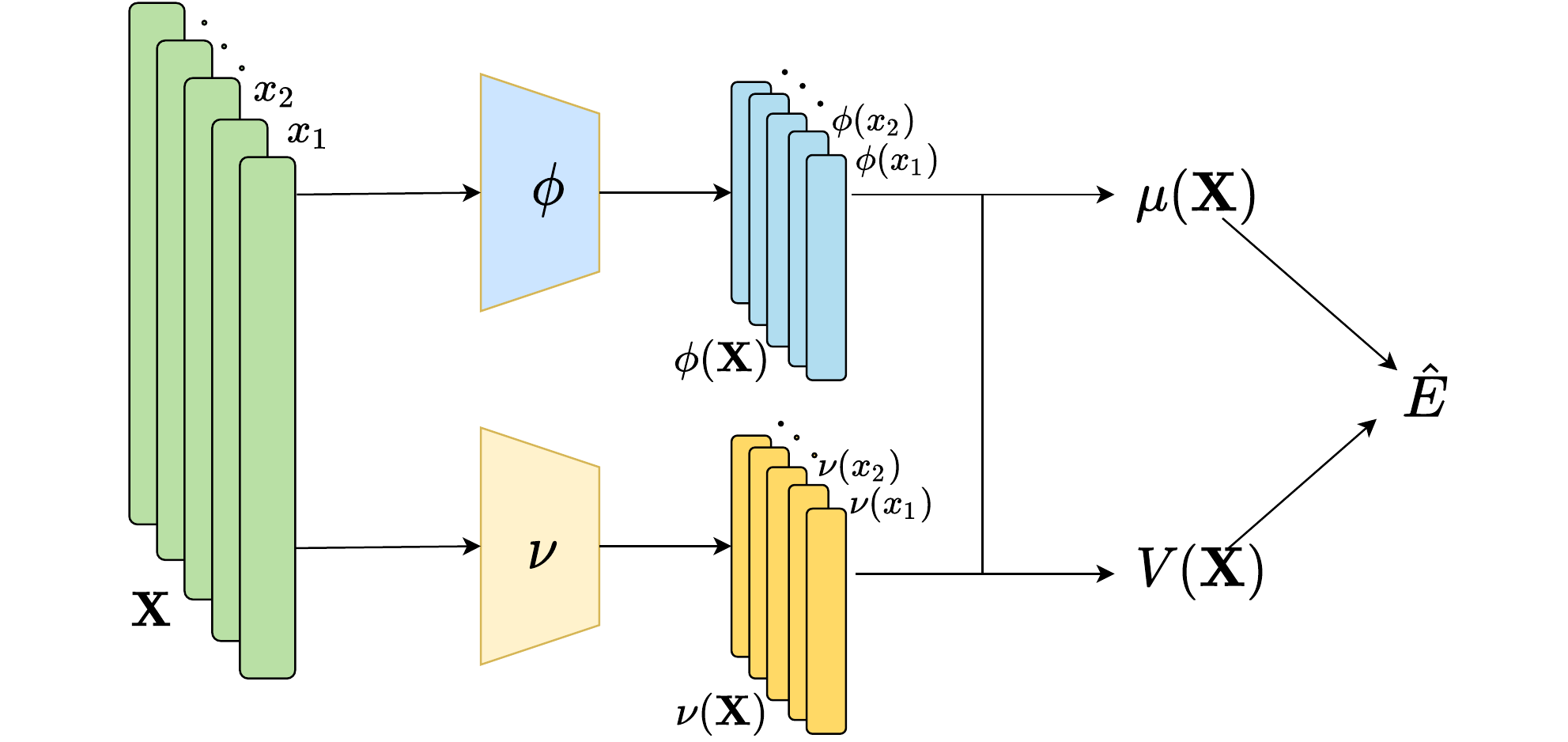}
	\caption[Domain Encoder]{The Domain Encoder Structure.}
	\label{encoder_b}
\end{figure}

An encoder that takes the whole dataset features and outputs an estimated environmental variable $E$ neglecting the permutation of sample orders for each specific domain is preferred in our case. According the theory of deep sets \citep{zaheer2017deep} below:
\begin{theorem}
	\citep{zaheer2017deep} A function $f(X)$ on a set $X$ having countable elements, is a valid set function, i.e. invariant to the permutation of objects in $X$ if and only if it can be decomposed as the form $\rho (\sum_{x\in X}\phi(x))$ for suitable transformations $\rho$ and $\phi$.
\end{theorem} 
The key to deep sets is to add up all representations and then apply nonlinear transformations. Further inspired by the approximated Bayesian posterior \citep{maeda2020meta} on the variable $E$, we design our encoder structure as shown in Figure~\ref{encoder_b} where:
\begin{align}
V(\mathbf{X}) &= (\sum_i^n\nu(x_i)-(n-1)\nu_0)^{-1}  \\
\mu(\mathbf{X}) &= V(\mathbf{X})(\sum_i^n\nu(x_i)\phi(x_i)).
\label{meanv}
\end{align}
For point estimation on $E$, we directly let $\hat{E} = \mu(\mathbf{X})$. For approximate Bayesian inference on $E$, we sample $\hat{E} \sim \mathcal{N}(\mu(\mathbf{X}),V(\mathbf{X}))$. See more about the intuition on encoder structure design in Appendix \ref{encoder}. 

\subsubsection{Structural Filters}
 
We directly use a weight matrix as each variable's structural filter, more details about which are shown in Figure \ref{DAPDAG}. As for other hidden layers in the decoder architecture, we keep them shared for all variables and these will be discussed in next sub-section. 

\subsubsection{Hidden and Output Layers}
\paragraph{Shared Hidden Layers}
The model is designed to have shared hidden layers out of two purposes: (1) Learning similar basis functions/representations among variables; (2) Saving the computation resource. 

As we have mentioned in assumptions, each variable follows a distribution of exponential family conditioned on its parents (and $E$). Since distributions in exponential family can be represented as a common form of probability density function, the shared hidden layers are expected to learn the similarity of basis representation among these variables that are assumed to follow conditional distributions from the same family.     

On the other hand, shared hidden layers can substantially reduce the efforts needed for computation during training the model. Normally, we would have separate hidden layers for each variable. However, this will introduce much more learning parameters, which decrease the model's scalability in high-dimensional setting and could also aggravate over-fitting facing small dataset. 

\paragraph{Separate Output Layers}
We have separate output layer for each variable of either a continuous type or binary type. For continuous variables, the output layer is simply a weight matrix without any activation function. For binary variables, the output layer will be a weight matrix with sigmoid activation function. 

\subsubsection{Loss Function} Denote $g, \Theta_1, \Theta_2,\Theta_3$ the parameters of encoder, structural filters, shared hidden layers and output layers respectively ($\Theta = \Theta_1 \cup \Theta_2 \cup \Theta_3 $), the model is trained by minimising the below loss function with respect to $g$ and $\Theta$ for each source domain index $m \in [M]$:
\begin{equation}
\label{loss}
\mathcal{L}_{m}= \mathcal{L}_{N_m}(\mathbf{Y}^m, f_{d+1}(g, \Theta)) + \gamma\hat{E}_m^2 + \lambda \mathcal{R}_{\mathcal{G}}(\mathbf{\tilde{X}}^m,f_{g, \Theta})
\end{equation} 
where for continuous variables:
\begin{equation}
\mathcal{L}_{N_m}(\mathbf{Y}^m, f_{d+1}(g, \Theta)) = \frac{1}{N_m}||\mathbf{Y}^m-f_{d+1}(\mathbf{X}^m)||^2 
\label{con-loss}
\end{equation}
and for binary variables:
\begin{align}
\mathcal{L}_{N_m}(\mathbf{Y}^m, f_{d+1}(g, \Theta)) = \frac{1}{N_m}\sum_{i=1}^{N_m}[\mathbf{Y}_i^m\log f_{d+1}(\mathbf{X}_i^m)\nonumber\\
+(1-\mathbf{Y}_i^m)\log (1-f_{d+1}(\mathbf{X}_i^m)]. 
\label{bin-loss}
\end{align} We also regularise the estimated $\hat{E}$ since a small $E$ is expected for better generalisation of decoder as shown in Theorem \ref{bound}. The DAG loss $\mathcal{R}_{\mathcal{G}}$ takes the form of:
\begin{align}
\label{dagloss}
\mathcal{R}_{\mathcal{G}}(\mathbf{\tilde{X}}^m,f_{g,\Theta})&=\mathcal{L}_{N_m}(f_{g, \Theta}(\mathbf{\tilde{X}}^m))+h(\Theta_1) \nonumber \\
&+\alpha h(\Theta_1)^2 +\beta l_1(\Theta_1).
\end{align}  
where $\mathcal{L}_{N_m}(f_{g, \Theta}(\mathbf{\tilde{X}}^m))$ is the reconstruction loss for all variables including features and the label in domain $m$. We use the mean squared loss (\ref{con-loss}) for continuous variables and cross entropy loss (\ref{bin-loss}) for binary variables. $h(\Theta_1)=0$ is the acyclicity constraint of NO-TEARS \citep{zheng2020learning}. $l_1(\Theta_1)$ is the group lasso regularisation on the weight matrix in $\Theta_1$. $\alpha$, $\beta$ and $\gamma$ are the corresponding hyper-parameters. 

\begin{figure}[h!]
	\centering
	\includegraphics[scale=0.5]{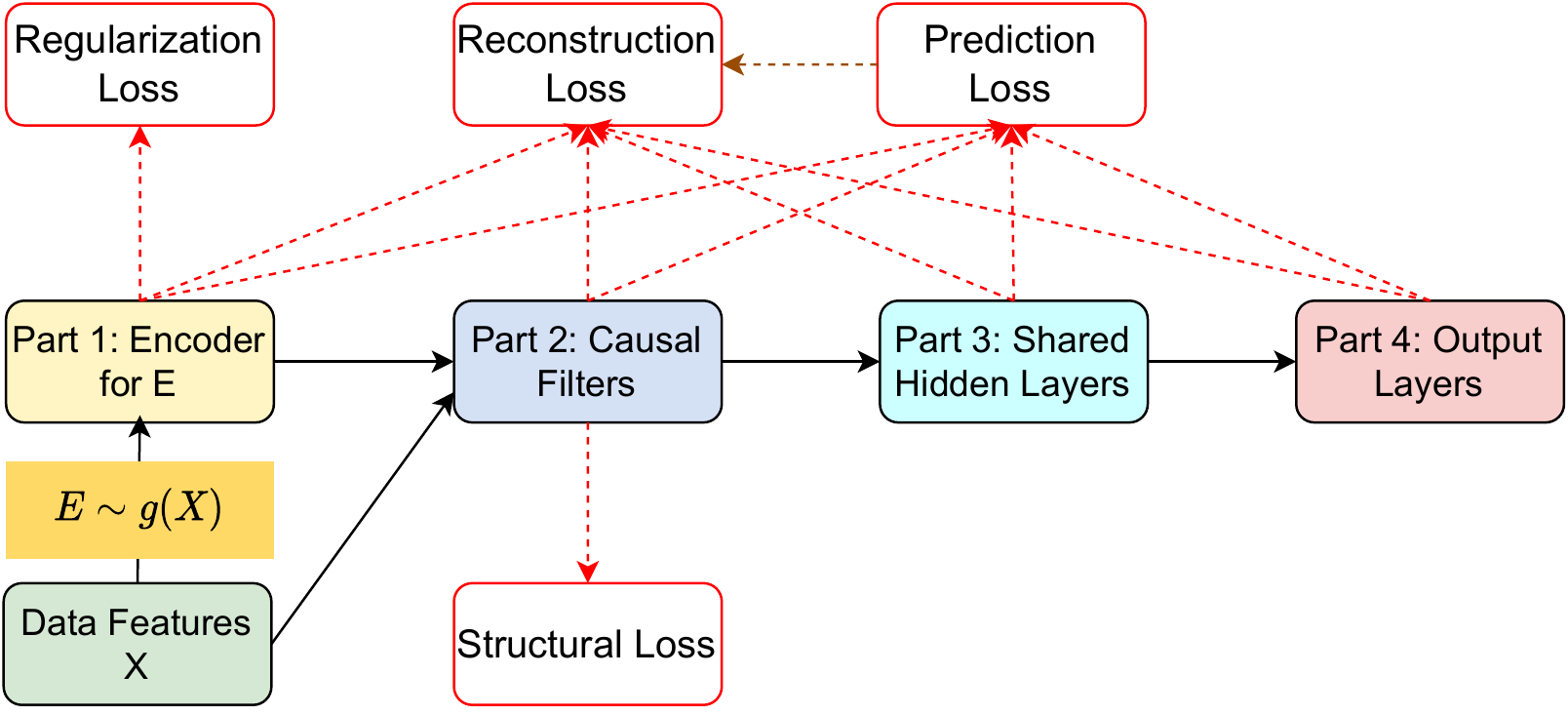}
	\caption[Loss Function]{Loss components and corresponding responsible model parts: the regularisation loss refers to square regularisation on $\hat{E}$ and the structural loss includes the DAG constraint and $l_1$ loss on structural filters.}
	\label{losscomp}
\end{figure}

\paragraph{Generalisation Bound of Decoder} We have derived a generalisation bound of the decoder $\Theta$ trained on i.i.d data within the same domain, which validates the form of our loss function (\ref{loss}).  
\begin{theorem}
	\label{bound}
	Let $f_{\Theta}$: $\tilde{\mathcal{X}} \rightarrow \tilde{\mathcal{X}}$ be a $L$-layer ReLU feed-forward neural network decoder with hidden layer size $h$. Then, under appropriate assumptions \ref{assum1}, \ref{assum2}, \ref{assum3} and \ref{assum4} on the neural network norm and loss functions (refer to Appendix \ref{assum} for more details), $\forall \delta \in (0,1)$, with probability at least $1-\delta$ on a training domain with $N$ i.i.d samples conditioned on a shared $E$, we have:
	\begin{align}
	\mathcal{L}_P(f_{\Theta}) &\leq 4\mathcal{L}_N^c(f_{\Theta}) + \mathcal{L}_N^b(f_{\Theta})  \nonumber \\
	& + \frac{3}{N}[\mathcal{R}_{\Theta_1}+C_1\cdot E^2 + C_2(\mathcal{V}(\Theta_1)+\mathcal{V}(\Theta_2)\nonumber \\
	&+ \mathcal{V}(\Theta_3) + \log(\frac{8}{\delta}))]+C_3
	\end{align}
\end{theorem}
where $C_1$, $C_2$ and $C_3$ are constants, $\mathcal{V}(\cdot)$ is the square of $l_2$ norm on the corresponding parameters and $\mathcal{R}_{\Theta_1}$ is the DAG constraint on $\Theta_1$. For more details on the theorem proof, please refer to Appendix \ref{generalization}.

\subsubsection{Training Strategy}
In this section, we introduce a novel algorithm for training our model with multiple domains. The flow chart of the training algorithm is depicted as in Figure~\ref{trainflow}. For more details, please refer to the Algorithm \ref{train} in supplementary materials \ref{algo}.    

\begin{figure}[ht!]
	\centering
	\includegraphics[scale=0.42]{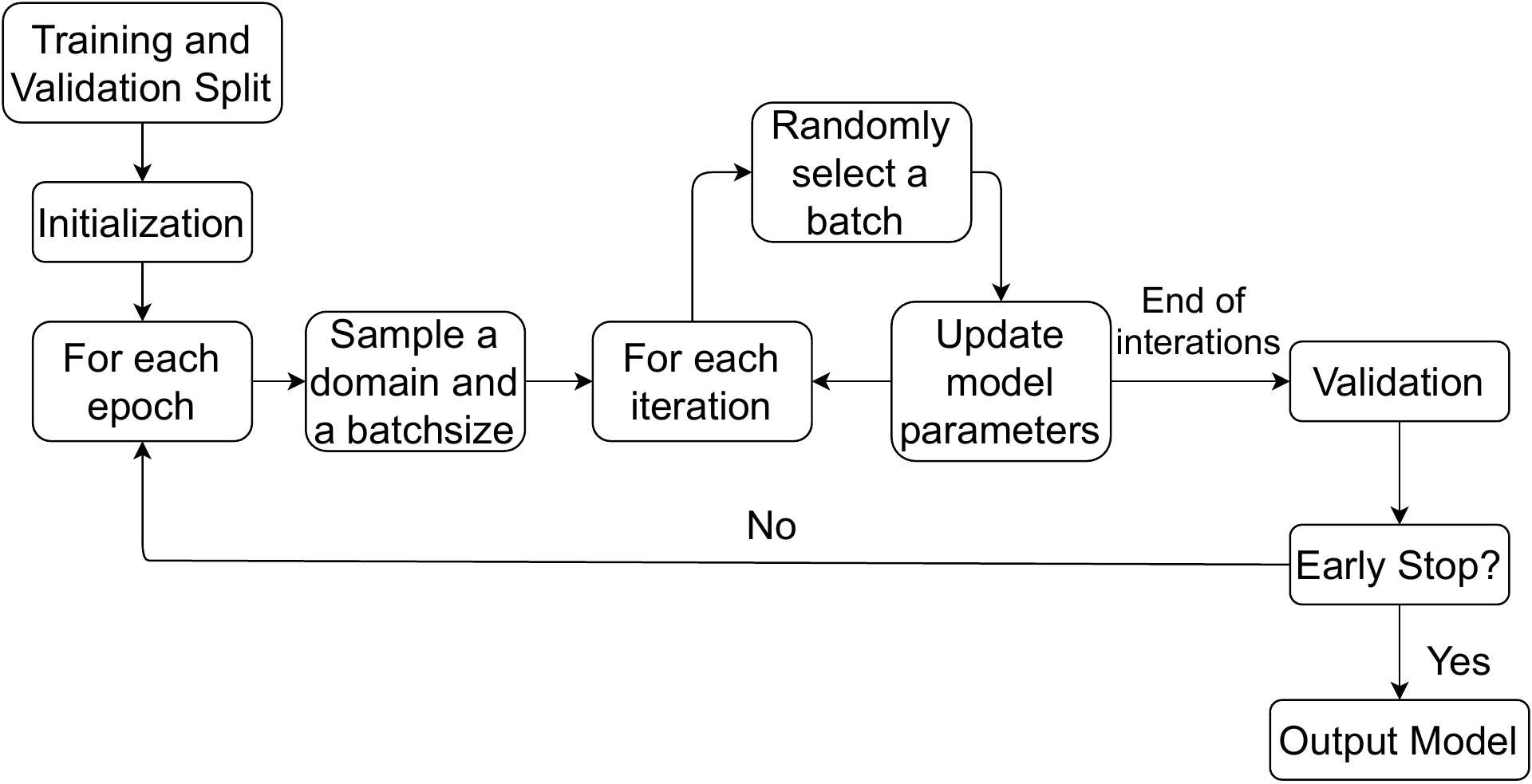}
	\caption[Training Strategy]{Training epochs: we iterate stochastic domain selection and updating parameters within each training epoch for N times where N is the integer (training size/sampled batch size). For each epoch, we randomly select a domain according to the size weights of source domains, and sample a batch size that is no larger than the size of selected training domain. Then we update the encoder and decoder parameters respectively with alternating maximisation-maximisation procedures. We use different batch sizes for different training epochs because we want to ensure the encoder learned will be applicable to both large and small domain datasets. After N iterations of each epoch, we validate the updated model using validation sets from all source domains. If the validation scores on all validation sets are not improving for a predefined patience number, we cease the training and output the model.}
	\label{trainflow}
\end{figure}

\paragraph{Prediction in the Target Domain}
To predict the target variable $Y^{\tau}$ in the target domain, we first feed features of the whole unlabelled dataset into the encoder to get the predicted $\hat{E}_{\tau}$. Then we go through corresponding model components by order: the last causal filter of $Y$, the hidden layers and the last output layer of $Y$ trained from source domains to get the predicted $\hat{Y}^{\tau}$ taking $\hat{E}_{\tau}$ and features $\mathbf{X}^{\tau}$ as input.

\subsubsection{Bayesian Formulation}
We can also put the whole framework into Bayesian formulation. The log likelihood of observed data $\mathbf{\tilde{X}}^m$ is
\begin{align}
\label{pd}
\log p(\mathbf{\tilde{X}}^m)
= &-\log\frac{q(E|\mathbf{X}^m)}{p(E)} + \log\frac{q(E|\mathbf{X}^m)}{p(E|\mathbf{\tilde{X}}^m)} \nonumber \\
& + \log p(\mathbf{\tilde{X}}^m|E) 
\end{align}

By taking the expectation on both sides of (\ref{pd}) with respect to a variational posterior $q(E|\mathbf{X}^m)$, the evidence lower bound (ELBO) of the marginal distribution of observed data is derived as below:
\begin{align}
\label{elbo}
\log p(\mathbf{\tilde{X}}^m)  \geq  & -KL(q(E|\mathbf{X}^m)||p(E)) \nonumber \\
&+ E_{q(E|\mathbf{X}^m)} [\sum_i^{n_m}\log p_{\Theta}(\mathbf{x}_i^m,y_i^m|E)]. 
\end{align} Where $KL(q(E|\mathbf{X}^m)||p(E))=\frac{1}{2}[-1+\log(\sigma_e^2)-\log(V(\mathbf{X}^m))+\frac{1}{\sigma_e^2}(\mu(\mathbf{X}^m)^2+V(\mathbf{X}^m))]$ if we assume $q(E|\mathbf{X}^m) \sim \mathcal{N}(\mu(\mathbf{X}^m),V(\mathbf{X}^m))$. It is easily noticed that this KL term also contains a squared regularisation term on estimated $\hat{E}$. We can then replace the prediction loss and reconstruction loss in (\ref{loss}) with corresponding ELBO to train the Bayesian predictor.
\paragraph{Prediction} After getting the trained decoder $\Theta$ and variational parameters $q_{g}$ (the encoder parameters), we perform prediction on the target domain $\tau$ by approximate inference via sampling
:
\begin{align}
P(y^{\tau}|\mathbf{x}^{\tau}) & = \int P(y^{\tau}|\mathbf{x}^{\tau}, E)q(E|\mathbf{X}^{\tau})d_E \nonumber \\
& \approx \frac{1}{N} \sum_{i=1}^N P(y^{\tau}|\mathbf{x}^{\tau},E_i)
\end{align}
where $E_i \sim q(E|\mathbf{X}^{\tau})$.

\section{Experiments}
In this section, we empirically evaluate the performance of our method for UDA on synthetic and real-world datasets. To begin with, we will briefly describe experiment settings including evaluation metrics, baselines and benchmarks we compare with. In the second sub-session, we discuss experiments on two made-up datasets which comply with our basic assumptions. We demonstrate the performance improvement of DAPDAG (our method) (Please refer to Appendix \ref{abl} for ablation studies on how each part of the model contributes to the performance gain). In the third section, we introduce real-world datasets - MAGGIC (Meta-Analysis Global Group in Chronic Heart Failure) \citep{martinez2012gender} with 30 different studies of patients and test our method on the processed datasets of selected studies against benchmarks.

\subsection{Experiment Setups}

\paragraph{Benchmarks} We benchmark DAPDAG against the plain MLP, CASTLE and MDAN (Multi-domain Adversarial Networks) \citep{zhao2018multiple} and BRM (Meta Learning as Bayesian Risk Minimisation) \citep{maeda2020meta}. We set MLP to be our baseline method and train it on merged data by directly combining all source domains. MDAN is representative of a class of well-founded DA methods \citep{pei2018multi,sebag2019multi} to learn an invariant feature representation or implicit distribution alignment across domains. They use an adversarial objective to minimise the training loss over labelled sources and distance of feature representation between each source domain and the target domain at the same time. Despite that this class of methods are usually applied in the field of computer vision with high-dimensional image data, we adapt the structure and transfer the idea to our learning setting where data are generated by a DAG with much fewer variables. While BRM can also be regarded as an auto-encoder that could make inference on latent variable perturbing the conditional distribution of $Y$ without reconstructing DAG as an auxiliary task.   

\paragraph{Implementation and Training} All methods are implemented using PyTorch driven by GPU. We set the same decoder architecture of DAPDAG as CASTLE except that DAPDAG has an extra domain encoder and an extra row for taking inferred $E$ in structural filters. Moreover, the DAPDAG decoder has the same number of hidden layers and number neurons in each hidden layer with MLP, BRM decoder and feature extractor of MDAN. We fix the number of hidden layers to be 2 and number of hidden neurons to be 16 for both synthetic and real datasets. For the encoder of DAPDAG and DoAMLP, we use a two-hidden-layer deep-set structure with the same number of neurons as decoder in each hidden layer. The activation function used is ELU and each model is trained using the Adam optimiser \cite{kingma2014adam} with an early stopping regime. For the data features with large scales in classification datasets such as ages, BMI (body-mass index), we standardise these variables with a mean of 0 and variance of 1.  

\subsection{Synthetic Datasets}
In this part, we present experiments on synthetic datasets, please refer to \ref{syn} in supplementary materials for more detailed description on synthetic datasets. 

\paragraph{Comparison with Benchmarks}
\begin{figure}[h!]
	\centering
	\includegraphics[scale=0.55]{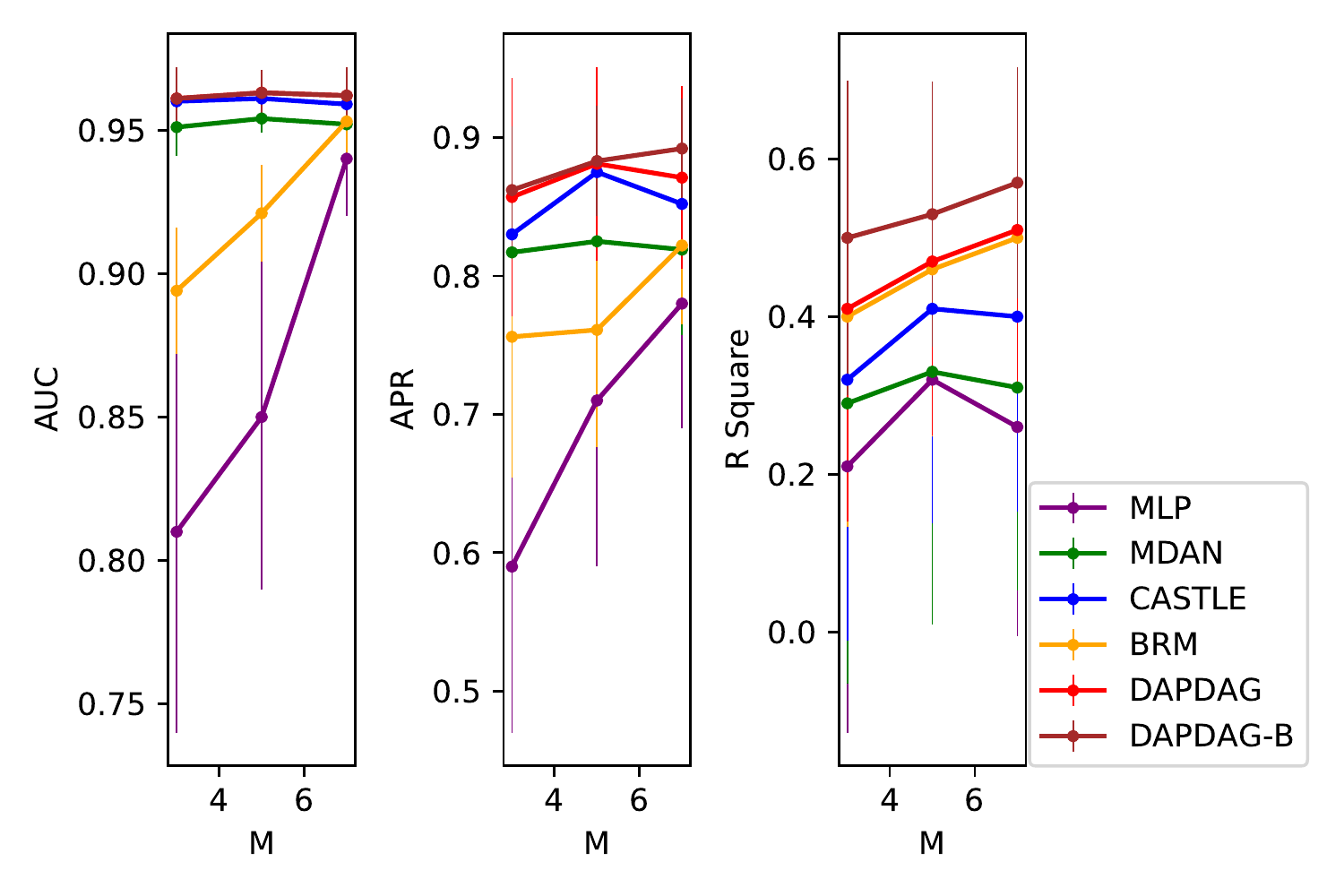}
	\caption{Performance against benchmarks on synthetic datasets. DAPDAG-B denotes the DAPDAG under Bayesian formulation.}
	\label{fig:syncomp}
\end{figure}

\begin{figure*}[h!]
	\centering
	\includegraphics[scale=0.44]{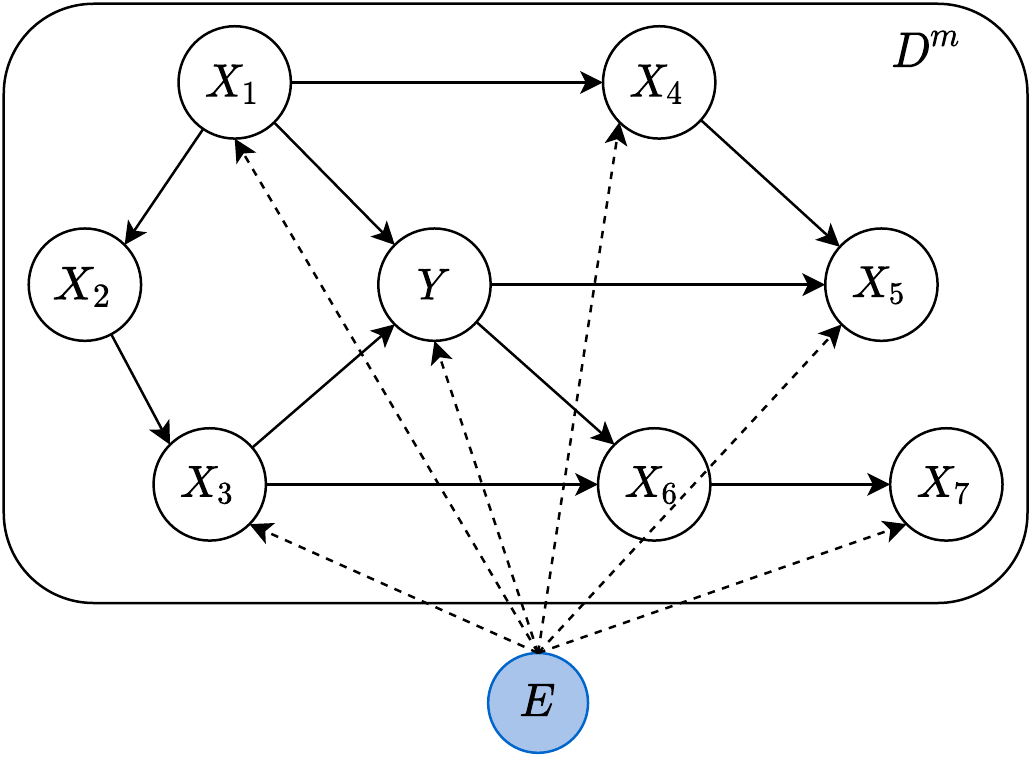}
	\includegraphics[scale=0.4]{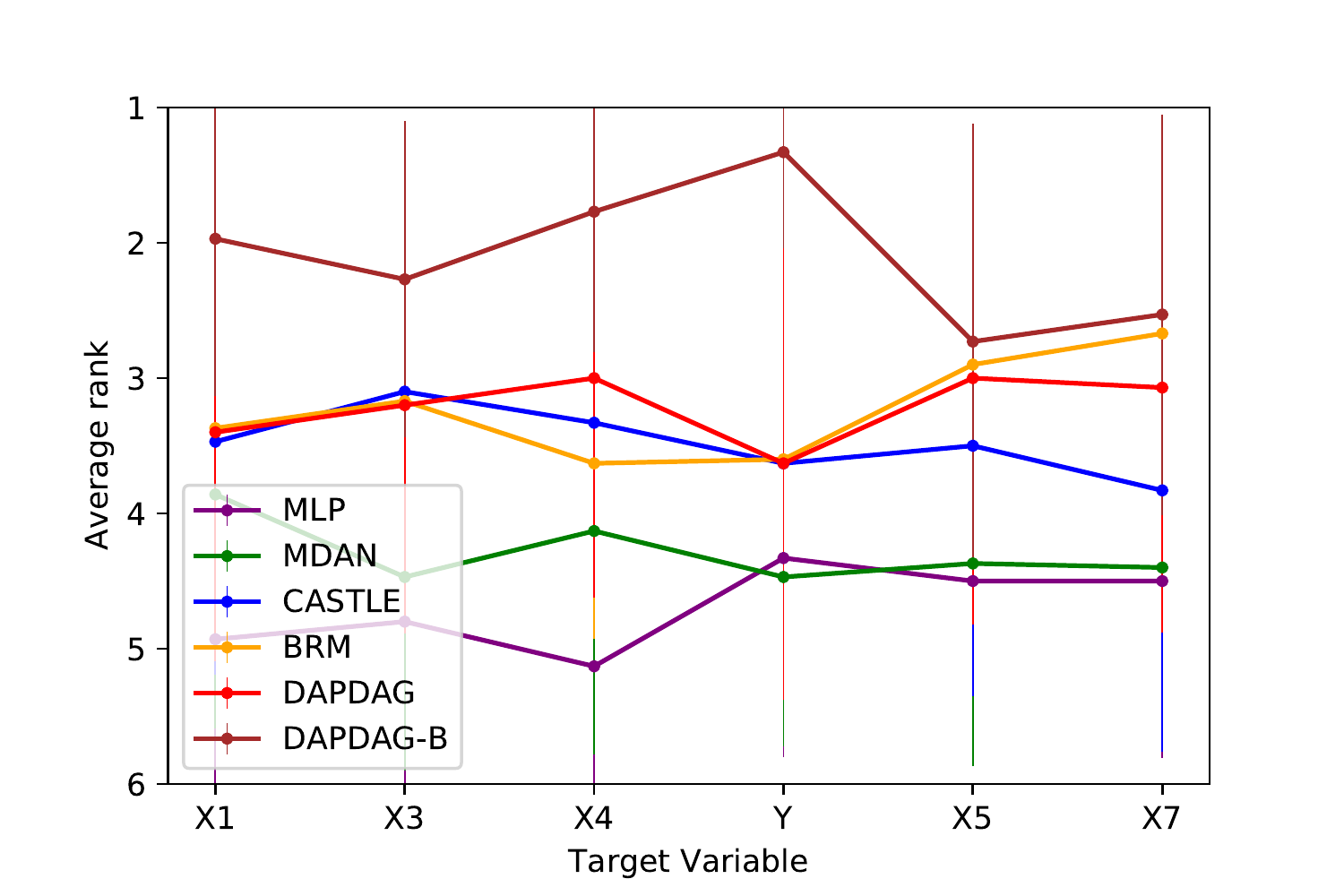}
	\includegraphics[scale=0.4]{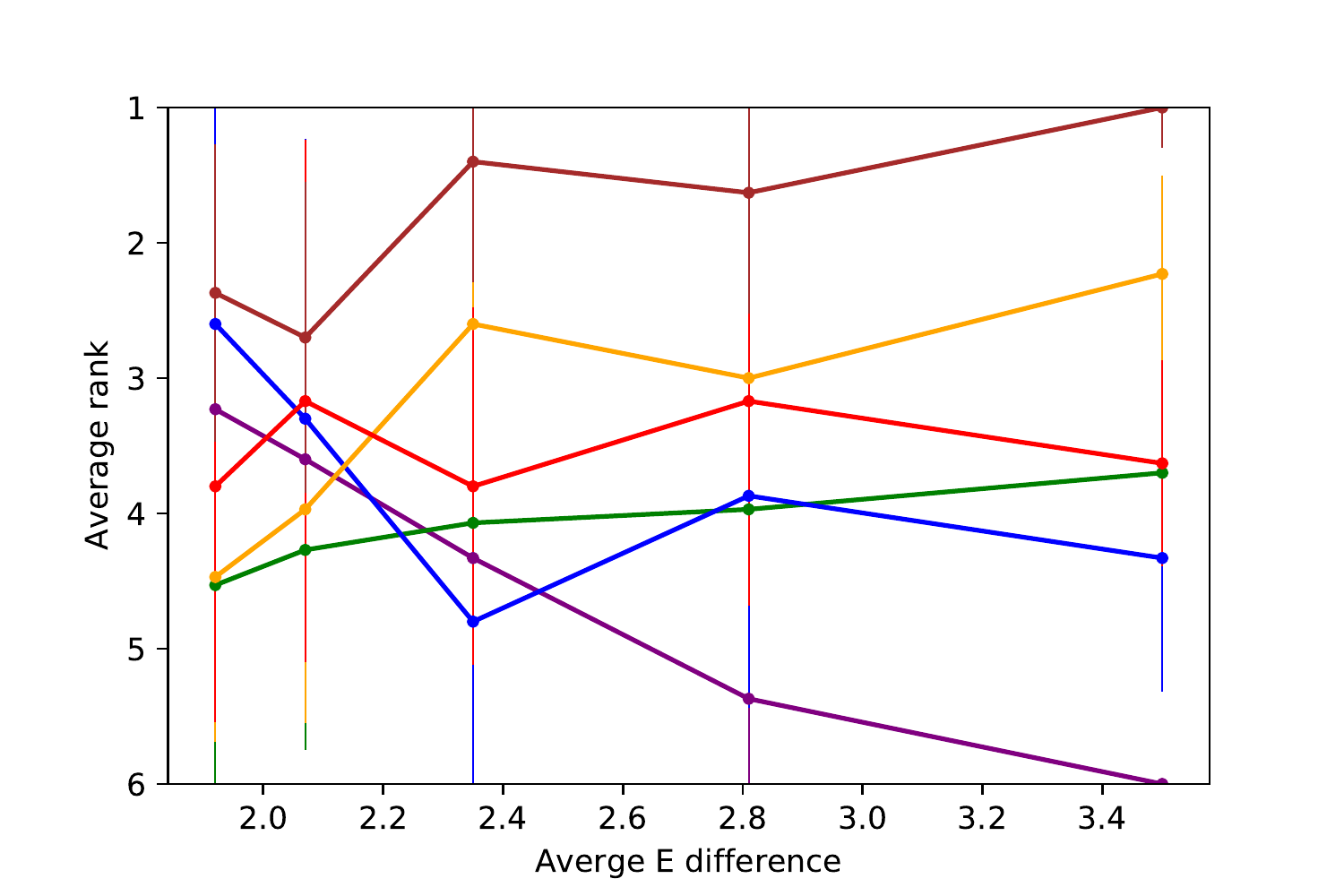}
	\label{rank-scores}
	\caption[Performance of DAPDAG against benchmarks on synthetic regression datasets]{Left sub-figure shows the DAG for generating the regression datasets (see more details in Appendix \ref{syn-gen}); Middle figure shows the average rank of each method's performance on the regression datasets with respect to different target variables selected from nodes in the left DAG. We repeat experiments for each target variable selection over 30 combinations of 9 sources and 1 target domain; Right figure plots the rank of each method's performance on the regression datasets with respect to the average distance between the target domain and source domains. Each averaged distance is the mean of absolute distance between the target domain and 9 sources.}  
\end{figure*}

We compare DAPDAG with other benchmarks with variant number of training sources with size of 500 for each domain set. As the results show in Figure~\ref{fig:syncomp}, DAPDAG outperforms all other benchmarks in both classification and regression datasets. Despite the fact that CASTLE does not have the ability to adjust for domain shift, it achieves better performance than MDAN with the ability of domain adaptation. This validates the intuition that in a causally perturbed system, forcing different distributions to be in a similar representation space may not help much compared to finding invariant causal features for prediction. However, these results only sketch a general performance gain of DAPDAG against other methods over multiple combinations of source and target domains. We also compare DAPDAG against benchmarks with respect to different target variables and average source-target difference. The results are shown in Figure~\ref{rank-scores}. We observe that DAPDAG has apparently better performance in scenarios where target domain is significantly different from sources and the target variable is not a sink node (that has no descendants) in the underlying causal DAG.  

\begin{figure}[h!]
    \centering
    \includegraphics[width=\linewidth]{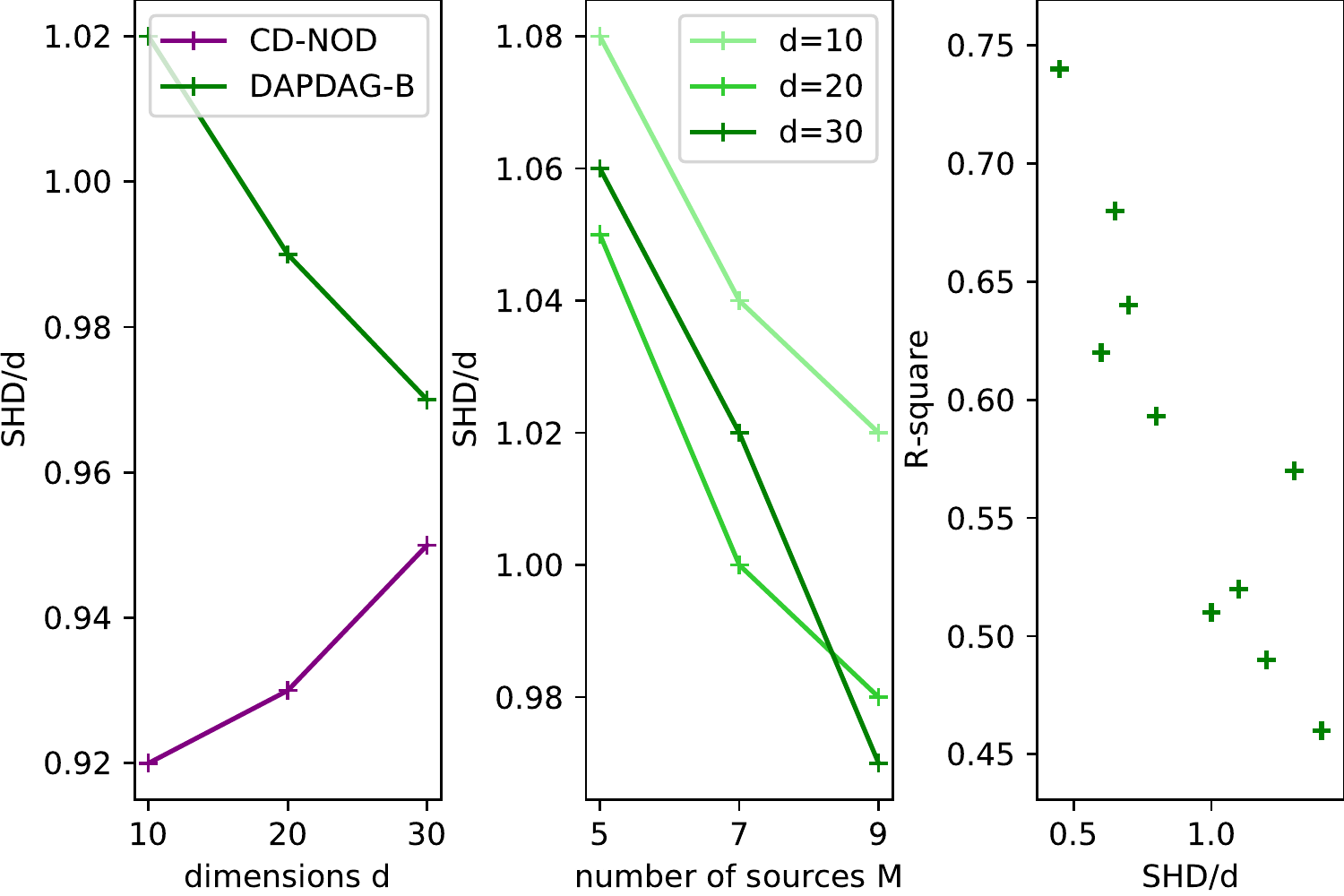}
    \caption{Evaluation on DAG Learning}
    \label{fig:dagl}
\end{figure}

\paragraph{Evaluation of DAG Learning} We have also included a few experiments evaluating the learned causal DAGs from synthetic regression datasets, as shown in Figure \ref{fig:dagl}, where $d$ is the number of variables, $M$ is the number of training domains and SHD is the structural hamming distance (used to measure the discrepancy between the learned graph and the truth, the lower the better). For generating synthetic datasets with different dimensions $d$, we randomly generate causal graphs and assign non-linear conditionals (For each variable $X_i = \mathbf{W}_i^2\sigma(\mathbf{W}_i^1[Pa_i])+\epsilon_i$ where both $\mathbf{W}_i^2$ and $\mathbf{W}_i^1$ are randomly sampled weight matrices, $Pa_i$ represent the graphical parents of the variable $i$, $\epsilon_i$ is the noise variable and $\sigma$ is the activation function) according to the causal order. We compare our method with the baseline CD-NOD \citep{zhang2017causal} in the left plot of Figure \ref{fig:dagl} (for non-oriented edges, we use the ground-truth directions if possible). Due to an extra prediction loss on the target variable in addition to the reconstruction loss, the learned graphs usually deviate from the truth in terms of mis-specified edges and redundant edges. Yet this prediction loss in the Bayesian formulation will become less important as dimensions increase and then the learned graph will approach the ground truth, which is shown in both the left plot and middle plot of the Figure \ref{fig:dagl}. Meanwhile, the right plot in Figure \ref{fig:dagl} demonstrates a highly positive relationship between the accuracy of graph learning and prediction performance.

\textbf{Scalability Analysis}: Please refer to part \ref{sca} in the appendix for more details.

\subsection{MAGGIC Datasets}
In this section, we show experimental results on MAGGIC dataset. Since DAPDAG-B (Bayesian formulation) performs better than DAPDAG on synthetic datasets, we only show the performance of DAPDAG in this part. We also add a benchmark - a data imputation method called MisForest \citep{stekhoven2012missforest} to impute labels in the target domain as missing values. MAGGIC is a collection of 30 datasets from different medical studies containing patients who experienced heart failure. For the UDA task, we take the 12 shared variables by all studies and set the label as one-year survival indicator.   

\begin{figure}[ht]
	\centering
	\includegraphics[scale=0.55]{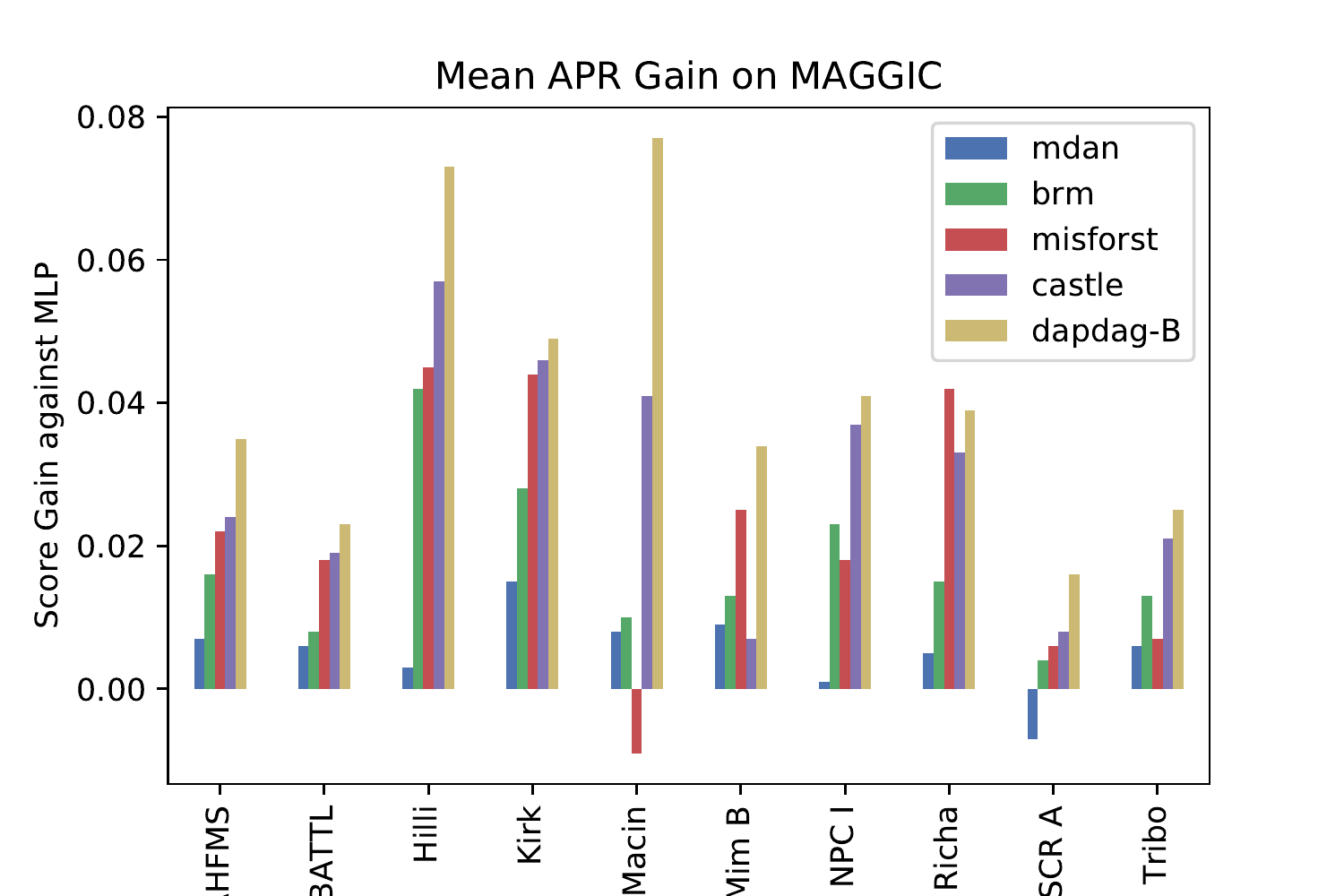}
	\label{maggic-scores}
	\caption[Performance of DAPDAG-B against benchmarks on MAGGIC datasets]{Performance of DAPDAG-B against benchmarks on MAGGIC datasets for each target study using rest studies as source domains.}
\end{figure} 

\paragraph{Performance} The experiment results on selected MAGGIC studies are demonstrated in Figure \ref{maggic-scores}. The shown results of our method are obtained using the environmental variable $\mathbf{E}$ with dimension 3, which is fine-tuned as a hyper-parameter during the model selection.We observe that DAPDAG-B can almost beat other benchmarks on the selected studies in APR scores. Despite the minor improvement against benchmarks in a few studies such as "BATTL" and "Kirk" or even worse performance than MissForest in "Richa", DAPDAG exhibits significant performance boost in other studies like "Hilli", "Macin" and "NPC I", which are found to be more different from rest sources (please refer to \ref{supp} in supplementary materials).

\section{Discussion}
\label{s:Summary}
To sum up, we explore a novel auto-encoder structure that combines estimation of population statistics using deep sets and reconstructing a DAG  through a regularised decoder. We prove that under certain assumptions, the loss function has components similar to terms in the generalisation bound of decoder, which validates the form of training loss. Experiments on synthetic and real datasets manifest the performance gain of our method against popular benchmarks in UDA tasks.    

\paragraph{Better design of encoder.} Currently, the encoder needs to take the whole dataset from a domain as input, which greatly slows down the training speed when the source dataset size is huge. Meanwhile, a source domain with large sample set is preferred since it will help capture the environmental variable $E$. Therefore, a better encoder should be designed to balance the trade-off from domain sizes.

\paragraph{Theoretical exploration on the encoder.} We have derived a generalisation bound for decoder within the same domain yet haven't looked into the properties of encoder. We hope to dive deeper into theoretical guarantees on the encoder for inference on $E$.

\paragraph{Extension to DA with Feature Mismatch.} Currently, we only focus on the task of UDA within the same feature space. In reality, it is highly possible to encounter datasets with different features available in each domain, such as the case of missing features across studies in MAGGIC dataset. Although imputation can be a solution, it can fail if there are a large portion of non-overlapped features for each domain. Therefore, it is imperative to develop approaches that can handle feature mismatch in the near future.

\newpage
\bibliography{example_paper}
\bibliographystyle{icml2022}

\newpage
\appendix
\onecolumn
\section{NOTEARS for Learning Non-linear SEM}
\label{dag-notears}
How to construct a proxy of $\mathbf{B}$ for a general non-linear SEM? Suppose in graph $\mathcal{G}$, there exists a function $f_i: \mathbb{R}^d \rightarrow \mathbb{R}$ for the $i$-th variable $X_i$ such that
\begin{equation}
\label{exp}
\mathbb{E}[X_i|X_{Pa(i)}] = f_i(\mathbf{X})
\end{equation}
where if $X_j \not\in Pa(i)$ then $f_i(x_1,...,x_{d+1})$ does not depend on $x_j$, leading to a result that the function $a(u) :  = f_i(x_1,...,x_{j-1},u,x_{j+1},...,x_{d+1})$ is constant for all $u\in \mathbb{R}$. \cite{zheng2020learning} uses partial derivatives $\frac{\partial f_i}{\partial x_j}$ to measure the dependence of $X_i$ on $X_j$.
Denote $\partial_j f_i = \frac{\partial f_i}{\partial x_j}$, then it can be shown that 
\begin{equation}
f_i \perp \!\!\! \perp X_j \Leftrightarrow ||\partial_j f_i||_{L^2} = 0
\end{equation}
where $||.||_{L^2}$ is the $L^2$-norm. Denote the matrix $\mathbf{A}(f) \in \mathbb{R}^{d\times d}$ with entries $[\mathbf{A}(f)]_{ij} := ||\partial_j f_i||_{L^2}$. Then $\mathbf{A}(f)$ becomes an non-linear surrogate of the adjacency matrix $\mathbf{B}$ in linear models. Now consider using a MLP to approximate the $f_i$. Suppose the MLP has $h$ hidden layers and a single activation $\sigma:\mathbb{R} \rightarrow \mathbb{R}$:
\begin{equation}
\hat{f}_i(\mathbf{u}) = \sigma(\sigma(...\sigma(\mathbf{u}\mathbf{W}_i^{(1)})\mathbf{W}_i^{(2)})\mathbf{W}_i^{(h)}),
\end{equation}
where $\mathbf{u} \in \mathbb{R}^d$ and $\mathbf{W}_i^{(l)} \in \mathbb{R}^{n_{l-1}\times n_l}$ and $n_0=d$. It is shown in \citep{zheng2020learning} that if $[\mathbf{W}_i^{(1)}]_{bk}=0$ for all $k=1,...,n_1$, then $\hat{f}_i(\mathbf{u}) $ is independent of the $k$-th input $u_k$. Let $\theta = (\theta_1,...,\theta_d)$ with $\theta_i = (\mathbf{W}_i^{(1)},...,\mathbf{W}_i^{(h)})$ and define $[A(\theta)]_{ij}$ as the norm of $j$-th row of $\mathbf{W}_i^{(1)}$. Then it suffices to solve DAG learning by tacking below problem \citep{zheng2020learning}:
\begin{equation}
\begin{aligned}
\min\limits_{\theta} \quad \frac{1}{n} \sum_{i=1}^{d} l(x_i,\hat{f}_i(\mathbf{X,\theta_i}))+\lambda ||\mathbf{W}_i^{(1)}||_{1,1} \\
\text{subject to} \quad h(A(\theta)) = 0
\end{aligned}
\end{equation}
\section{Intuition on the Encoder Design}
\label{encoder}
In this part, we intuitively induce the encoder design drawn from Bayesian posterior of $E$. Our objective is to infer the latent variable $E$ from a sample of features. Following the idea of Bayesian inference, the MAP estimate of a latent variable can be obtained by maximising its posterior. In our case, however, we aim to learn a direct but approximate mapping from the features to the key statistics of $E$ posterior distribution given those features if its posterior is assumed to have a special form of distribution, e.g. Gaussian.

Consider the observed data $\{\mathbf{\tilde{x}}^m\}_{i=1}^{N_m}$ in source domain $m\in [M]$. For notation simplicity, we omit the domain index $m$ in following texts. let's begin with the conditional probability of $\{\mathbf{\tilde{x}}\}_{i=1}^{n}$ i.i.d data drawn from the same domain, we have:

\begin{equation}
p(\{\mathbf{\tilde{x}}\}_{i=1}^{n}|E) = \prod_{i=1}^{n}p(\mathbf{\tilde{x}}_i|E). 
\end{equation}
For the posterior of $E$ given $\{\mathbf{\tilde{x}}\}_{i=1}^{n}$, we have:
\begin{align}
p(E|\{\mathbf{\tilde{x}}\}_{i=1}^{n}) &= \frac{p(\{\mathbf{\tilde{x}}\}_{i=1}^{n}|E)\cdot p(E)}{p(\{\mathbf{\tilde{x}}\}_{i=1}^{n})} \nonumber \\
&\propto p(\{\mathbf{\tilde{x}}\}_{i=1}^{n}|E)\cdot p(E) \nonumber \\
&\propto p(E)\prod_{i=1}^{n}p(\mathbf{\tilde{x}}_i|E) \nonumber \\
&\propto p(E)\prod_{i=1}^{n}\frac{p(E|\mathbf{\tilde{x}}_i)}{p(E)} \nonumber \\
&\propto p(E)^{-(n-1)}\prod_{i=1}^{n}p(E|\mathbf{\tilde{x}}_i).
\end{align}
If we further assume both $p(E)$ and $p(E|\mathbf{\tilde{x}}_i)$ are members of an exponential family, e.g. Gaussian distributions (without loss of generality), which can be expressed (approximately) as:
\begin{align}
   p(E) &= \mathcal{N}(0, \nu_0^{-1}) \\
   p(E|\mathbf{\tilde{x}}_i) &=  \mathcal{N}(\phi(\mathbf{\tilde{x}}_i), \nu^{-1}(\mathbf{\tilde{x}}_i))
\end{align}
where $\phi$, $\nu^{-1}$ are approximated mappings and $\nu_0^{-1}$ is the parameter for the prior variance of $E$. Then we can re-write $p(E|\{\mathbf{\tilde{x}}\}_{i=1}^{n})$ as:
\begin{align}
p(E|\{\mathbf{\tilde{x}}\}_{i=1}^{n}) &\propto \exp(-0.5(1-n)\nu_0\cdot E^2)\prod_{i=1}^{n}\exp(-0.5\nu(\mathbf{\tilde{x}}_i)\cdot (E-\phi(\mathbf{\tilde{x}}_i))^2)  \nonumber \\
&\propto \exp(-0.5[(1-n)\nu_0 E^2+\sum_{i=1}^n\nu(\mathbf{\tilde{x}}_i)\cdot (E-\phi(\mathbf{\tilde{x}}_i))^2])  \nonumber \\
&\propto \exp(-0.5[((1-n)\nu_0+\sum_{i=1}^n\nu(\mathbf{\tilde{x}}_i))E^2-2(\sum_{i=1}^n\phi(\mathbf{\tilde{x}}_i)\nu(\mathbf{\tilde{x}}_i))E])
\label{posterior}
\end{align}
By completion of squares on (\ref{posterior}), we get the approximate posterior $p(E|\{\mathbf{\tilde{x}}\}_{i=1}^{n}) \sim \mathcal{N}(\mu(\mathbf{\tilde{X}}),V(\mathbf{\tilde{X}}))$ with the similar form as in (\ref{meanv}) except that the input is $\mathbf{\tilde{X}}$ in (\ref{posterior}) instead of $\mathbf{X}$ in (\ref{meanv}). 

\section{Generalisation Bound with Mixed Type Data}
\label{generalization} 

Our proof of Theorem \ref{bound} mainly follows the work by \cite{kyono2020castle}, except the extension to mixed data type including binary variables and regularisation on the environmental variable $E$. Let $\mathcal{L}_P(f_{\Theta})$ and $\mathcal{L}_N(f_{\Theta})$ be the expected loss and empirical loss respectively. We further divide each loss into two components - $\mathcal{L}^c(f_{\Theta})$ as the loss of continuous variables and $\mathcal{L}^b(f_{\Theta})$ as the loss of binary variables. Similar notations of distinguishing variable types are applied to $\Theta$, $f_{\Theta}$ and $\tilde{\mathbf{X}}_i$.

\subsection{Assumptions}
\label{assum}
\begin{assumption}
	\label{assum1}
	For any sample $\tilde{\mathbf{X}} = (\mathbf{X},Y) \sim P_{\tilde{\mathbf{X}}} $, the continuous variables $\tilde{\mathbf{X}}^c$ has bounded $l_2$ norm such that $\exists B_1 > 0, ||\tilde{\mathbf{X}}^c||_2 \leq B_1$. This can further infer that \cite{kyono2020castle}:
	\begin{equation}
	\sup\limits_{\tilde{\mathbf{X}}\in \tilde{\mathcal{X}}} \mathbb{E}_{\mathbf{u}}||f_{\Theta_{\mathbf{u}}}^c(\tilde{\mathbf{X}})-f_{\Theta}^c(\tilde{\mathbf{X}})||^2 \leq \gamma_1
	\end{equation} where $\gamma_1$ is a constant. 
\end{assumption}

\begin{assumption}
	\label{assum2}
	For any sample $\tilde{\mathbf{X}} = (\mathbf{X},Y) \sim P_{\tilde{\mathbf{X}}} $, we assume
	\begin{equation}
	\sup\limits_{\tilde{\mathbf{X}}\in \tilde{\mathcal{X}}} \max \{\sum_{j=1}^{b}\mathbb{E}_{\mathbf{u}}||\log \frac{f_{\Theta}^{b_j}(\tilde{\mathbf{X}})}{f_{\Theta_{\mathbf{u}}}^{b_j}(\tilde{\mathbf{X}})}||, \sum_{j=1}^{b}\mathbb{E}_{\mathbf{u}}||\log \frac{1-f_{\Theta}^{b_j}(\tilde{\mathbf{X}})}{1-f_{\Theta_{\mathbf{u}}}^{b_j}(\tilde{\mathbf{X}})}||\}  \leq \gamma_2
	\end{equation} where $\gamma_2$ is a constant. 
\end{assumption}

\begin{assumption}
	\label{assum3}
	The squared loss function of continuous variables $\mathcal{L}^c(f_{\Theta}) = ||f_{\Theta}^c(\tilde{\mathbf{X}})-\tilde{\mathbf{X}}^c||^2$ is sub-Gaussian under the distribution $P_{\tilde{\mathbf{X}}}$ with a proxy-variance factor $s_1^2$ such that $\forall \epsilon \in \mathbb{R}$, $ \mathbb{E}_P[\exp(\epsilon(\mathcal{L}^c(f_{\Theta})-\mathcal{L}_P^c(f_{\Theta})))] \leq \exp(\frac{\epsilon^2s_1^2}{2})$.    
\end{assumption}

\begin{assumption}
	\label{assum4}
	The loss function for binary variables $\mathcal{L}^b(f_{\Theta}) =  \sum_{j=1}^{b}\mathcal{L}^{b_j}(f_{\Theta})$ where $\mathcal{L}^{b_j}$ is the cross-entropy loss function of $j$-th binary variable, is sub-Gaussian under the distribution $P_{\tilde{\mathbf{X}}}$ with a proxy-variance factor $s_2^2$ such that $\forall \epsilon \in \mathbb{R}$, $ \mathbb{E}_P[\exp(\epsilon(\mathcal{L}^b(f_{\Theta})-\mathcal{L}_P^b(f_{\Theta})))] \leq \exp(\frac{\epsilon^2s_2^2}{2})$.  
\end{assumption}

\subsection{Generalisation Bound for the Decoder}
\textit{Proof}. Denote $\Theta$ as the parameters of DAPDAG decoder and $\Theta_{\mathbf{u}}$ as perturbed $\Theta$ where each parameter in $\Theta$ is perturbed by a noise vector $\mathbf{u} \sim \mathcal{N}(\mathbf{0},\sigma^2\mathbf{I})$.  

\paragraph{Step 1.} We first the derive the upper bound for the expected loss over parameter perturbation and data distribution. For each shared $E$ within the same domain, we have (for simplicity, we omit the notation of $E$, which serves as an input for $f_{\Theta}$, $f_{\Theta_{\mathbf{u}}}$, $\mathcal{L}(f_{\Theta})$ and $\mathcal{L}(f_{\Theta_{\mathbf{u}}})$ ):
\begin{align}
\label{lnc}
\mathbb{E}_{\mathbf{u}}[\mathcal{L}_N^c(f_{\Theta_{\mathbf{u}}})] & = \mathbb{E}_{\mathbf{u}}[\frac{1}{N}\sum_{i=1}^N||f_{\Theta_{\mathbf{u}}}^c(\tilde{\mathbf{X}}_i)-f_{\Theta}^c(\tilde{\mathbf{X}}_i)+f_{\Theta}^c(\tilde{\mathbf{X}}_i)-\tilde{\mathbf{X}}_i^c||^2] \nonumber \\
& = \mathbb{E}_{\mathbf{u}}[\frac{1}{N}\sum_{i=1}^N||f_{\Theta_{\mathbf{u}}}^c(\tilde{\mathbf{X}}_i)-f_{\Theta}^c(\tilde{\mathbf{X}}_i)||^2]+\frac{1}{N}\sum_{i=1}^N||f_{\Theta}^c(\tilde{\mathbf{X}}_i)-\tilde{\mathbf{X}}_i^c||^2 \nonumber \\
& + \mathbb{E}_{\mathbf{u}}[\frac{2}{N}\sum_{i=1}^N(f_{\Theta_{\mathbf{u}}}^c(\tilde{\mathbf{X}}_i)-f_{\Theta}^c(\tilde{\mathbf{X}}_i))\cdot (f_{\Theta}^c(\tilde{\mathbf{X}}_i)-\tilde{\mathbf{X}}_i^c)] \nonumber \\
& \leq 
\mathbb{E}_{\mathbf{u}}[\frac{1}{N}\sum_{i=1}^N||f_{\Theta_{\mathbf{u}}}^c(\tilde{\mathbf{X}}_i)-f_{\Theta}^c(\tilde{\mathbf{X}}_i)||^2]+\frac{1}{N}\sum_{i=1}^N||f_{\Theta}^c(\tilde{\mathbf{X}}_i)-\tilde{\mathbf{X}}_i^c||^2 \nonumber \\
& + \mathbb{E}_{\mathbf{u}}[\frac{1}{N}\sum_{i=1}^N||f_{\Theta_{\mathbf{u}}}^c(\tilde{\mathbf{X}}_i)-f_{\Theta}^c(\tilde{\mathbf{X}}_i)||^2]+\frac{1}{N}\sum_{i=1}^N||f_{\Theta}^c(\tilde{\mathbf{X}}_i)-\tilde{\mathbf{X}}_i^c||^2 \nonumber \\
& \leq 
2 \gamma_1 + 2 \mathcal{L}_N^c(f_{\Theta})
\end{align}
Similarly, we can derive:
\begin{align}
\label{lpc}
\mathcal{L}_P^c(f_{\Theta}) & = \mathbb{E}_P\mathbb{E}_{\mathbf{u}}||f_{\Theta}^c(\tilde{\mathbf{X}})-f_{\Theta_{\mathbf{u}}}^c(\tilde{\mathbf{X}})+f_{\Theta_{\mathbf{u}}}^c(\tilde{\mathbf{X}})-\tilde{\mathbf{X}}^c||^2 \nonumber \\
& \leq 
2 \gamma_1 + 2\mathbb{E}_{\mathbf{u}}[\mathcal{L}_P^c(f_{\Theta_{\mathbf{u}}})] \nonumber \\
\end{align}
where $\gamma_1$ is the upper bound of $\mathbb{E}_{\mathbf{u}}||f_{\Theta_{\mathbf{u}}}^c(\tilde{\mathbf{X}})-f_{\Theta}^c(\tilde{\mathbf{X}})||^2$ such that $ \sup\limits_{\tilde{\mathbf{X}}\in \tilde{\mathcal{X}}} \mathbb{E}_{\mathbf{u}}||f_{\Theta_{\mathbf{u}}}^c(\tilde{\mathbf{X}})-f_{\Theta}^c(\tilde{\mathbf{X}})||^2 \leq \gamma_1$. Let $Q_{\Theta_{\mathbf{u}}}$ and $P_{\Theta_{\mathbf{u}}}$ be the distribution and prior distribution of perturbed decoder parameter $\Theta_{\mathbf{u}}$, $Q_E$ and $P_E$ be the distribution and prior distribution of $E$  respectively. According to Corollary 4 in \cite{germain2016pac} and original proof of CASTLE, we can trivially transfer their theoretical results to continuous variables in our framework. Given $P_{\Theta_{\mathbf{u}}}$ and $P_E$ that are independent of training data in the domain with $E$, we can deduce from the PAC-Bayes theorem that with probability at least $1-\delta$  $\forall \delta \in (0,1)$, for any $N$ i.i.d training samples with the shared $E$:
\begin{align}
\label{pac-con}
\mathbb{E}_{\mathbf{u}}[\mathcal{L}_P^c(f_{\Theta_{\mathbf{u}}})] & \leq \mathbb{E}_{\mathbf{u}}[\mathcal{L}_N^c(f_{\Theta_{\mathbf{u}}})] + \frac{1}{N}[2 KL(Q_{E,\Theta_{\mathbf{u}}^c}||P_{E,\Theta_{\mathbf{u}}^c}) + \log\frac{8}{\delta}] + \frac{s_1^2}{2}
\end{align} 
where $Q_{E,\Theta_{\mathbf{u}}^c} = Q_{\Theta_{\mathbf{u}}^c} \cdot Q_{E}$ and $P_{E,\Theta_{\mathbf{u}}^c} = P_{\Theta_{\mathbf{u}}^c} \cdot P_{E}$. Combining \ref{lnc}, \ref{lpc} and \ref{pac-con}, we get:
\begin{align}
\label{pac-con2}
\mathcal{L}_P^c(f_{\Theta}) & \leq 
6 \gamma_1 + 4\mathcal{L}_N^c(f_{\Theta}) + \frac{2}{N}[2 KL(Q_{E,\Theta_{\mathbf{u}}^c}||P_{E,\Theta_{\mathbf{u}}^c}) + \log\frac{8}{\delta}] + s_1^2 
\end{align}

For $j$-th binary variable denoted as $\tilde{\mathbf{X}}^{b_j}$, we have:
\begin{align}
\mathbb{E}_{\mathbf{u}}[\mathcal{L}_N^{b_j}(f_{\Theta_{\mathbf{u}}})] & = -\mathbb{E}_{\mathbf{u}}[\frac{1}{N}\sum_{i=1}^N(\tilde{\mathbf{X}}_i^{b_j}\log f_{\Theta_{\mathbf{u}}}^{b_j}(\tilde{\mathbf{X}}_i)+(1-\tilde{\mathbf{X}}_i^{b_j})\log (1-f_{\Theta_{\mathbf{u}}}^{b_j}(\tilde{\mathbf{X}}_i) ))] \nonumber \\
& + \mathbb{E}_{\mathbf{u}}[\frac{1}{N}\sum_{i=1}^N(\tilde{\mathbf{X}}_i^{b_j}\log f_{\Theta}^{b_j}(\tilde{\mathbf{X}}_i)+(1-\tilde{\mathbf{X}}_i^{b_j})\log (1-f_{\Theta}^{b_j}(\tilde{\mathbf{X}}_i) ))] \nonumber \\
& - \mathbb{E}_{\mathbf{u}}[\frac{1}{N}\sum_{i=1}^N(\tilde{\mathbf{X}}_i^{b_j}\log f_{\Theta}^{b_j}(\tilde{\mathbf{X}}_i)+(1-\tilde{\mathbf{X}}_i^{b_j})\log (1-f_{\Theta}^{b_j}(\tilde{\mathbf{X}}_i) ))] \nonumber \\
& = -\mathbb{E}_{\mathbf{u}}[\frac{1}{N}\sum_{i=1}^N(\tilde{\mathbf{X}}_i^{b_j}\log \frac{f_{\Theta_{\mathbf{u}}}^{b_j}(\tilde{\mathbf{X}}_i)}{f_{\Theta}^{b_j}(\tilde{\mathbf{X}}_i)} +(1-\tilde{\mathbf{X}}_i^{b_j})\log \frac{1-f_{\Theta_{\mathbf{u}}}^{b_j}(\tilde{\mathbf{X}}_i)}{1-f_{\Theta}^{b_j}(\tilde{\mathbf{X}}_i)})] + \mathcal{L}_N^{b_j}(f_{\Theta}) \nonumber \\
& = \frac{1}{N}\sum_{i=1}^N\mathbb{E}_{\mathbf{u}}[\tilde{\mathbf{X}}_i^{b_j}\log \frac{f_{\Theta}^{b_j}(\tilde{\mathbf{X}}_i)}{f_{\Theta_{\mathbf{u}}}^{b_j}(\tilde{\mathbf{X}}_i)} +(1-\tilde{\mathbf{X}}_i^{b_j})\log \frac{1-f_{\Theta}^{b_j}(\tilde{\mathbf{X}}_i)}{1-f_{\Theta_{\mathbf{u}}}^{b_j}(\tilde{\mathbf{X}}_i)}] + \mathcal{L}_N^{b_j}(f_{\Theta}) \nonumber \\
& \leq \frac{1}{N}\sum_{i=1}^N\mathbb{E}_{\mathbf{u}}[||\tilde{\mathbf{X}}_i^{b_j}\log \frac{f_{\Theta}^{b_j}(\tilde{\mathbf{X}}_i)}{f_{\Theta_{\mathbf{u}}}^{b_j}(\tilde{\mathbf{X}}_i)}|| + ||(1-\tilde{\mathbf{X}}_i^{b_j})\log \frac{1-f_{\Theta}^{b_j}(\tilde{\mathbf{X}}_i)}{1-f_{\Theta_{\mathbf{u}}}^{b_j}(\tilde{\mathbf{X}}_i)}||] + \mathcal{L}_N^{b_j}(f_{\Theta}) \nonumber \\
& \leq \frac{1}{N}\sum_{i=1}^N\mathbb{E}_{\mathbf{u}}[||\log \frac{f_{\Theta}^{b_j}(\tilde{\mathbf{X}}_i)}{f_{\Theta_{\mathbf{u}}}^{b_j}(\tilde{\mathbf{X}}_i)}|| + ||\log \frac{1-f_{\Theta}^{b_j}(\tilde{\mathbf{X}}_i)}{1-f_{\Theta_{\mathbf{u}}}^{b_j}(\tilde{\mathbf{X}}_i)}||] + \mathcal{L}_N^{b_j}(f_{\Theta})
\end{align}
We then upper bound the expected perturbed loss for all binary variables:
\begin{align}
\mathbb{E}_{\mathbf{u}}[\mathcal{L}_N^{b}(f_{\Theta_{\mathbf{u}}})] & = \mathbb{E}_{\mathbf{u}}[\sum_{j=1}^{b}\mathcal{L}_N^{b_j}(f_{\Theta_{\mathbf{u}}})] \nonumber \\
& \leq \sum_{j=1}^{b}  \frac{1}{N}\sum_{i=1}^N\mathbb{E}_{\mathbf{u}}[||\log \frac{f_{\Theta}^{b_j}(\tilde{\mathbf{X}}_i)}{f_{\Theta_{\mathbf{u}}}^{b_j}(\tilde{\mathbf{X}}_i)}|| + ||\log \frac{1-f_{\Theta}^{b_j}(\tilde{\mathbf{X}}_i)}{1-f_{\Theta_{\mathbf{u}}}^{b_j}(\tilde{\mathbf{X}}_i)}||] + \sum_{j=1}^{b} \mathcal{L}_N^{b_j}(f_{\Theta}) \nonumber \\
& = \frac{1}{N}\sum_{i=1}^N[\sum_{j=1}^{b}\mathbb{E}_{\mathbf{u}} ||\log \frac{f_{\Theta}^{b_j}(\tilde{\mathbf{X}}_i)}{f_{\Theta_{\mathbf{u}}}^{b_j}(\tilde{\mathbf{X}}_i)}|| + \sum_{j=1}^{b}\mathbb{E}_{\mathbf{u}}||\log \frac{1-f_{\Theta}^{b_j}(\tilde{\mathbf{X}}_i)}{1-f_{\Theta_{\mathbf{u}}}^{b_j}(\tilde{\mathbf{X}}_i)}||] + \mathcal{L}_N^{b}(f_{\Theta}) \nonumber \\
& \leq 2\gamma_2 + \mathcal{L}_N^{b}(f_{\Theta})
\end{align}
where $\gamma_2$ is a constant such that $ \sup\limits_{\tilde{\mathbf{X}}\in \tilde{\mathcal{X}}} \max \{\sum_{j=1}^{b}\mathbb{E}_{\mathbf{u}}||\log \frac{f_{\Theta}^{b_j}(\tilde{\mathbf{X}})}{f_{\Theta_{\mathbf{u}}}^{b_j}(\tilde{\mathbf{X}})}||, \sum_{j=1}^{b}\mathbb{E}_{\mathbf{u}}||\log \frac{1-f_{\Theta}^{b_j}(\tilde{\mathbf{X}})}{1-f_{\Theta_{\mathbf{u}}}^{b_j}(\tilde{\mathbf{X}})}||\}  \leq \gamma_2$. Similar to \ref{lpc}, we also have:
\begin{align}
\mathcal{L}_P^b(f_{\Theta}) &= \mathbb{E}_P\mathbb{E}_{\mathbf{u}} [\sum_{j=1}^b(\tilde{\mathbf{X}}^{b_j}\log f_{\Theta}^{b_j}(\tilde{\mathbf{X}})+(1-\tilde{\mathbf{X}}^{b_j})\log (1-f_{\Theta}^{b_j}(\tilde{\mathbf{X}}) ))] \nonumber \\
& - \mathbb{E}_P\mathbb{E}_{\mathbf{u}}[\sum_{j=1}^b(\tilde{\mathbf{X}}^{b_j}\log f_{\Theta_{\mathbf{u}}}^{b_j}(\tilde{\mathbf{X}})+(1-\tilde{\mathbf{X}}^{b_j})\log (1-f_{\Theta_{\mathbf{u}}}^{b_j}(\tilde{\mathbf{X}})))] \nonumber \\
& + \mathbb{E}_P\mathbb{E}_{\mathbf{u}}[\sum_{j=1}^b(\tilde{\mathbf{X}}^{b_j}\log f_{\Theta_{\mathbf{u}}}^{b_j}(\tilde{\mathbf{X}})+(1-\tilde{\mathbf{X}}^{b_j})\log (1-f_{\Theta_{\mathbf{u}}}^{b_j}(\tilde{\mathbf{X}}) ))] \nonumber \\
&  \leq 2\gamma_2 + \mathbb{E}_{\mathbf{u}}[\mathcal{L}_P^b(f_{\Theta_{\mathbf{u}}})] 
\end{align}
Similar to \ref{pac-con}, given $P_{\Theta_{\mathbf{u}}}$ and $P_E$ that are independent of training data in the domain with $E$, we can deduce from the PAC-Bayes theorem that with probability at least $1-\delta$  $\forall \delta \in (0,1)$, for any $N$ i.i.d training samples with the shared $E$:
\begin{align}
\label{pac-b}
\mathbb{E}_{\mathbf{u}}[\mathcal{L}_P^b(f_{\Theta_{\mathbf{u}}})] & \leq \mathbb{E}_{\mathbf{u}}[\mathcal{L}_N^b(f_{\Theta_{\mathbf{u}}})] + \frac{1}{N}[2 KL(Q_{E,\Theta_{\mathbf{u}}^b}||P_{E,\Theta_{\mathbf{u}}^b}) + \log\frac{8}{\delta}] + \frac{s_2^2}{2}
\end{align} 
where $Q_{E,\Theta_{\mathbf{u}}^b} = Q_{\Theta_{\mathbf{u}}^b} \cdot Q_{E}$ and $P_{E,\Theta_{\mathbf{u}}^b} = P_{\Theta_{\mathbf{u}}^b} \cdot P_{E}$. Combining results from \ref{lnc}, \ref{lpc} and \ref{pac-con}, we get:
\begin{align}
\label{pac-b2}
\mathcal{L}_P^b(f_{\Theta}) & \leq 
4 \gamma_2 + \mathcal{L}_N^b(f_{\Theta}) + \frac{1}{N}[2 KL(Q_{E,\Theta_{\mathbf{u}}^b}||P_{E,\Theta_{\mathbf{u}}^b}) + \log\frac{8}{\delta}] + \frac{s_2^2}{2}
\end{align}

\paragraph{Step 2.} Notice that $\Theta = \Theta_1 \cup \Theta_2 \cup \Theta_3 $ where $\Theta_1$, $\Theta_2$ and $\Theta_3$ represent the parameter of structural filters, shared hidden layers and output layers, we can further dissemble $\Theta_i$ for $i=1,2,3$ and write $P_{\Theta_{\mathbf{u}}}$ and $Q_{\Theta_{\mathbf{u}}}$ in more details. Let $\mathbf{W}$ be the weight matrix in a neural network layer and $L$ be the number of hidden layers, then we denote:
\begin{equation}
\Theta_1^c = \{\mathbf{W}_1^{c_j}\}_{j=1}^{c}, \quad \Theta_2^c = \{\mathbf{W}_k\}_{k=2}^{L}, \quad \Theta_3^c = \{\mathbf{W}_o^{c_j}\}_{j=1}^{c}
\end{equation}
as the decoder parameters of continuous variables. And the similar denotation for binary variables are as below:
\begin{equation}
\Theta_1^b = \{\mathbf{W}_1^{b_j}\}_{j=1}^{b}, \quad \Theta_2^b = \{\mathbf{W}_k\}_{k=2}^{L}, \quad \Theta_3^b = \{\mathbf{W}_o^{b_j}\}_{j=1}^{b}.
\end{equation}
Therefore, it is obvious that:
\begin{equation}
\Theta_1 = \Theta_1^c \cup \Theta_1^b = \{\mathbf{W}_1^{j}\}_{j=1}^{d+1}, \quad \Theta_2 = \Theta_2^c = \Theta_2^b = \{\mathbf{W}_k\}_{k=2}^{L}, \quad \Theta_3 = \Theta_3^c \cup \Theta_3^b = \{W_o^{j}\}_{j=1}^{d+1}
\end{equation}
Furthermore, we assume both $P_{\Theta_{\mathbf{u}}}$ and $Q_{\Theta_{\mathbf{u}}}$ can be decomposed into two parts such that:
\begin{equation}
P_{\Theta_{\mathbf{u}}} = P_{\Theta_{\mathbf{u}}}^1 \cdot P_{\Theta_{\mathbf{u}}}^2, \quad  Q_{\Theta_{\mathbf{u}}} = P_{\Theta_{\mathbf{u}}}^1 \cdot Q_{\Theta_{\mathbf{u}}}^2
\end{equation}
where $P_{\Theta_{\mathbf{u}}}^1$ and $Q_{\Theta_{\mathbf{u}}}^1$ are corresponding prior and probability distributions of structural filters $\Theta_1$ that form a DAG, $P_{\Theta_{\mathbf{u}}}^2$ and $Q_{\Theta_{\mathbf{u}}}^2$ are weight parameter distributions of corresponding layer parameters. Without loss of generality, $P_{\Theta_{\mathbf{u}}}^1$ and $Q_{\Theta_{\mathbf{u}}}^1$ are assumed to follow normal distributions for simplicity:
\begin{equation}
P_{\Theta_{\mathbf{u}}}^1 = P_{\Theta_{1,\mathbf{u}}}^1 \sim \mathcal{N}(h_{\Theta_{1,\mathbf{u}}}; d+1, 1), \quad Q_{\Theta_{\mathbf{u}}}^1 = Q_{\Theta_{1,\mathbf{u}}}^1 \sim \mathcal{N}(h_{\Theta_{1,\mathbf{u}}}; h_{\Theta_{1}}, 1) \\
\end{equation}
and the variable $h_{\Theta_{1,\mathbf{u}}}$ and constant $h_{\Theta_1}$ take the form as:
\begin{equation}
\label{hnorm}
h_{\Theta_{1,\mathbf{u}}} = Tr(\exp(\mathbf{A}_{\mathbf{u}}\odot \mathbf{A}_{\mathbf{u}})), \quad h_{\Theta_1} = Tr(\exp(\mathbf{A}\odot \mathbf{A}))
\end{equation}
where $\mathbf{A}_{\mathbf{u}}$ is a $(d+1)\times(d+1)$ adjacency-proxy matrix such that $[\mathbf{A}_{\mathbf{u}}]_{i,j}$ is the $l_2$-norm of the $i$-th row of the $j$-th perturbed structural filter matrix $W_{1,\mathbf{u}}^{j}$ and $\odot$ represents the Hadamard product operation. From the introduction of NOTEARS method before, we know that $h_{\Theta_{1,\mathbf{u}}} = Tr(\mathbf{I}) + \sum_{k=1}^{\infty}\frac{1}{k!}\sum_{i=1}^{d+1}[(\mathbf{A}_{\mathbf{u}}\odot \mathbf{A}_{\mathbf{u}})^k]_{ii} \geq d+1 $ and in fact each element in $\mathbf{A}_{\mathbf{u}}$ is non-negative, so using Normal approximation in \ref{hnorm} may not be appropriate for Bayesian Inference. Formally, it is better to consider using truncated normal or exponential priors for better approximation. 

And $P_{\Theta_{\mathbf{u}}}^2$ and $Q_{\Theta_{\mathbf{u}}}^2$ are given as:
\begin{align}
P_{\Theta_{\mathbf{u}}}^2 &= \prod_{j=1}^{d+1}\mathcal{N}(\mathbf{W}_{1,\mathbf{u}}^j;\mathbf{0},\sigma^2\mathbf{I})\prod_{k=2}^{L}\mathcal{N}(\mathbf{W}_{k,\mathbf{u}};\mathbf{0},\sigma^2\mathbf{I})\prod_{j=1}^{d+1}\mathcal{N}(\mathbf{W}_{o,\mathbf{u}}^j;\mathbf{0},\sigma^2\mathbf{I}), \\
Q_{\Theta_{\mathbf{u}}}^2 &= \prod_{j=1}^{d+1}\mathcal{N}(\mathbf{W}_{1,\mathbf{u}}^j;\mathbf{W}_{1}^j,\sigma^2\mathbf{I})\prod_{k=2}^{L}\mathcal{N}(\mathbf{W}_{k,\mathbf{u}};\mathbf{W}_{k},\sigma^2\mathbf{I})\prod_{j=1}^{d+1}\mathcal{N}(\mathbf{W}_{o,\mathbf{u}}^j;\mathbf{W}_{o}^j,\sigma^2\mathbf{I}).
\end{align}    
Recall that we also have a shared environmental variable $E$, which can be considered as a parameter independent of each component in the decoder. Despite that the $E$ value is obtained from the encoder taking sample features as input, for any $N$ i.i.d samples drawn from the same domain, this $E$ is fixed as a constant for decoder. Here, we further assume:
\begin{equation}
P_E = \mathcal{N}(0, \sigma_e^2), \quad Q_E= \mathcal{N}(E, \sigma_e^2).
\end{equation} 

\paragraph{Step 3.} 
By using the fact the that KL of two joint distributions is greater or equal to the KL of two marginal distributions, we can upper bound the KL in \ref{pac-con2} and \ref{pac-b2} using their versions of joint distributions:
\begin{align}
KL(Q_{E,\Theta_{\mathbf{u}}^b}||P_{E,\Theta_{\mathbf{u}}^b}) \leq KL(Q_{E,\Theta_{\mathbf{u}}}||P_{E,\Theta_{\mathbf{u}}}), \quad KL(Q_{E,\Theta_{\mathbf{u}}^c}||P_{E,\Theta_{\mathbf{u}}^c}) \leq KL(Q_{E,\Theta_{\mathbf{u}}}||P_{E,\Theta_{\mathbf{u}}}).
\end{align}
And we can upper bound  $KL(Q_{E,\Theta_{\mathbf{u}}}||P_{E,\Theta_{\mathbf{u}}})$ as follows:
\begin{align}
\label{kl}
KL(Q_{E,\Theta_{\mathbf{u}}}||P_{E,\Theta_{\mathbf{u}}}) &= \int Q_{E,\Theta_{\mathbf{u}}} \log\frac{Q_{E,\Theta_{\mathbf{u}}}}{P_{E,\Theta_{\mathbf{u}}}} d_E d_{\Theta_{\mathbf{u}}}  \nonumber \\
&= \int Q_E Q_{\Theta_{\mathbf{u}}}^1 Q_{\Theta_{\mathbf{u}}}^2 \log\frac{Q_E Q_{\Theta_{\mathbf{u}}}^1 Q_{\Theta_{\mathbf{u}}}^2}{P_E P_{\Theta_{\mathbf{u}}}^1 P_{\Theta_{\mathbf{u}}}^2} d_E d_{\Theta_{\mathbf{u}}}  \nonumber \\
&= \int Q_E Q_{\Theta_{\mathbf{u}}}^1 Q_{\Theta_{\mathbf{u}}}^2 \log\frac{Q_E}{P_E} d_E d_{\Theta_{\mathbf{u}}} + \int Q_E Q_{\Theta_{\mathbf{u}}}^1 Q_{\Theta_{\mathbf{u}}}^2 \log\frac{ Q_{\Theta_{\mathbf{u}}}^1}{P_{\Theta_{\mathbf{u}}}^1 } d_E d_{\Theta_{\mathbf{u}}} \nonumber \\
& +\int Q_E Q_{\Theta_{\mathbf{u}}}^1 Q_{\Theta_{\mathbf{u}}}^2 \log\frac{Q_{\Theta_{\mathbf{u}}}^2}{ P_{\Theta_{\mathbf{u}}}^2} d_E d_{\Theta_{\mathbf{u}}} \nonumber \\
& \leq \int Q_E \log\frac{Q_E}{P_E} d_E + \int Q_{\Theta_{\mathbf{u}}}^1 \log\frac{ Q_{\Theta_{\mathbf{u}}}^1}{P_{\Theta_{\mathbf{u}}}^1 } d_{\Theta_{\mathbf{u}}}  +\int Q_{\Theta_{\mathbf{u}}}^2 \log\frac{Q_{\Theta_{\mathbf{u}}}^2}{ P_{\Theta_{\mathbf{u}}}^2} d_{\Theta_{\mathbf{u}}} \nonumber \\
& = \frac{E^2}{2\sigma_e^2} + \frac{1}{2}[h_{\Theta_{1}}-(d+1)]^2 + \frac{1}{2\sigma^2}[\sum_{j=1}^{d+1}||\mathbf{W}_1^j||_F^2+\sum_{k=1}^{L}||\mathbf{W}_k||_F^2+\sum_{j=1}^{d+1}||\mathbf{W}_o^j||_F^2].
\end{align}
By upper-bounding the \ref{pac-b2} and \ref{pac-con2} using \ref{kl}, we have the final generalisation bound of decoder for mixed-type variables. Given $P_{\Theta_{\mathbf{u}}}$ and $P_E$ that are independent of training data within each domain, for any $N$ i.i.d training samples with the shared $E$, then with probability at least $1-\delta$  $\forall \delta \in (0,1)$ we have: 
\begin{align}
\label{pac-bound}
\mathcal{L}_P(f_{\Theta},E) & = \mathcal{L}_P^c(f_{\Theta},E) +\mathcal{L}_P^b(f_{\Theta},E) \nonumber \\
&\leq 
4\mathcal{L}_N^c(f_{\Theta}) + \mathcal{L}_N^b(f_{\Theta}) + \frac{3}{N}[\frac{E^2}{\sigma_e^2} + [h_{\Theta_{1}}-(d+1)]^2 \nonumber \\ 
&+ \frac{1}{\sigma^2}(\sum_{j=1}^{d+1}||\mathbf{W}_1^j||_F^2+\sum_{k=1}^{L}||\mathbf{W}_k||_F^2+\sum_{j=1}^{d+1}||\mathbf{W}_o^j||_F^2) + \log\frac{8}{\delta}] + C_3 
\end{align}
where $C_3 = 6 \gamma_1 + 4\gamma_2+ s_1^2 + \frac{s_2^2}{2}$.

\section{Training Algorithm}
\label{algo}
This section looks into more details about the training algorithm of DAPDAG. The sudo-code of the algorithm is shown in Algorithm \ref{train}.
\begin{algorithm}[tb]
	\caption{Training Algorithm for DAPDAG}
	\label{train}
\begin{algorithmic}
   \STATE {\bfseries Input:} $(\mathbf{X}_i^m,Y_i^m)_{i=1}^{n_m}$ for $m \in [M]$ where $M$ is the number of source domains; validation ratio $p$; patience $k$ for early stop.
    \STATE {\bfseries Output:} Domain Encoder $\phi$; structural filters $\Theta_1$; Shared hidden layers $\Theta_2$; Output layers $\Theta_3$ (decoder $\Theta=\Theta_1\bigcup\Theta_2\bigcup\Theta_3$).	
	\FOR{source index $m \in [M]$}
		\STATE Randomly split $(\mathbf{X}_i^m,Y_i^m)_{i=1}^{n_m}$ into training and validation datasets according to $p$;
		\STATE Record the size of training data: $N_m $;
	\ENDFOR	
	\STATE Obtain number of total training samples from all domains: $N=\sum_{m=1}^M N_m$;
	\FOR{source index $m \in [M]$}
	\STATE Compute the weight for each training domain: $w_m= \frac{N_m}{N}$; 
	\ENDFOR
	
	\STATE Initialise all parameters; 
	\FOR{each training epoch}
		\FOR{ index $i \in [M]$}
			\STATE Randomly select a training domain $m \sim Cat(w_1,w_2,...,w_M)$;
			\STATE Obtain the objective \ref{elbo} for the selected domain $(\mathbf{X}_i^m,Y_i^m)_{i=1}^{N_m}$;
			\STATE Update encoder parameters $\phi$ by maximising \ref{elbo} with respect to $\phi$;
			\STATE Update decoder parameters $\Theta$ by maximising \ref{elbo} with respect to $\Theta$.
		\ENDFOR
		\STATE Compute sum of validation scores from all validation sets;
		\IF{validation score not improving for k epochs}
			\STATE break the epoch.
		\ENDIF
	\ENDFOR
\end{algorithmic}
	
\end{algorithm}

\section{Experiments}
\subsection{Metrics} 
\paragraph{Classification} For classification task, we report two scores: Area Under ROC Curve (AUC) and Average Precision-Recall Score (APR). An ROC curve (receiver operating characteristic curve) plots True Positive Rate (TPR) versus False Positive Rate (FPR) at different classification thresholds, showing the performance of a classifier in a more balanced and robust manner. APR summarises a precision-recall curve as the weighted mean of precision attained at each pre-defined threshold, with the increase in recall from the previous threshold used as the weight:
\begin{equation}
APR = \sum_{n}(R_n-R_{n-1})P_n
\end{equation}
where $P_n$ and $R_n$ are the precision and recall at the $n$-th threshold. Both AUC and APR are computed using the predicted probabilities from classifier and the true labels in binary classification. 

\paragraph{Regression} For regression task, we present the coefficient of determination ($R^2$), the proportion of the variation in $Y$ that is predictable from $\mathbf{X}$.

\subsection{Benchmark: Domain-invariant Representation Methods}
\begin{figure}[h!]
	\centering
	\includegraphics[scale=0.4]{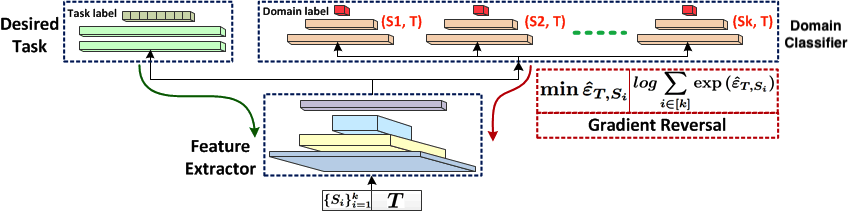}
	\caption[The Architecture of MDAN]{The illustration of MDAN (extracted from original paper): $\{S_i\}_i^{k}$ and $T$ are indices of source domains and the target domain respectively. The model has a shared multi-layer feature extractor (just same as hidden layers in a plain MLP). The extracted feature vector is then used to reconstruct the label, against which the training loss is minimised over multiple source domains. In the meantime, the feature vector of an instance in source domain $S_i$ is also used to fool the specific domain classifier that intends to distinguish feature vectors from $S_i$ and the target domain $T$.}
	\label{mdan}
\end{figure}

Here we give a more detailed description on adversarial methods for UDA with implicit alignment. Please refer to Figure \ref{mdan} for a general idea of the class of methods.

We have also added the data-driven unsupervised domain adaptation proposed by \cite{zhang2020domain} in extra comparison experiments. Because it requires a two-stage learning and much more parameters than our approach, we do not include it in the main texts. For more details, please see the Figure \ref{fig:sca} for the experiments on synthetic regression datasets.

\subsection{Synthetic Dataset Generation}
\label{syn}
We make two synthetic datasets for classification and regression task respectively: the classification dataset is made up following a DAG learned from MAGGIC dataset and the regression dataset is generated by our own DAG design in Figure \ref{syndag}. The general algorithm of synthetic generation is exhibited in Algorithm  \ref{syn-gen}.

\begin{algorithm}[h!]
	\caption{Generation Algorithm for Synthetic Datasets}
	\label{syn-gen}
    \begin{algorithmic}
    \STATE {\bfseries Input:} Random seed for sampling; Number of domains $M$; Required hyper-parameters $N$ and $\sigma^2$.
	\STATE {\bfseries Output:} Synthetic Datasets.	
	\FOR{ $m \in [M]$}
		\STATE Sample an environmental variable $\mathbf{E}_m \sim \mathcal{N}(0,\sigma^2)$;
		\STATE Sample a domain size $N_m \sim Pois(N)$;
		\FOR{ $i \in [N_m]$}
			\STATE Generate classification data according to \ref{cla-syn}; 
			\STATE Generate regression data according to \ref{reg-syn}. 
			\ENDFOR
	\ENDFOR
    \end{algorithmic}
\end{algorithm}

\subsubsection{Classification} 

We refer to the learned causal graph in \cite{kyono2019improving} as our ground truth for synthetic classification data (as shown in the right part of Figure \ref{syndag}). The made-up dataset have features that carry explicit meaning in real world thus they are generated compatible with reality to some extent (e.g. design of variable types, range of values, positive and negative causal relations should acknowledge the real-world constraints such as ages can not be negative.). We use 8 features to predict $Y$: 5-year survival rate of "made-up" patients. These features are $X_1$: Age of patients; $X_2$: Ethnicity of the patient; $X_3$: Angina; $X_4$: Myocardial Infarction; $X_5$: ACE Inhibitors; $X_6$: NYHA1; $X_7$: NYHA2; $X_8$: NYHA3. Equations \ref{cla-syn} below elaborate more details about their distributions and causal relationships. 
\begin{equation}
\label{cla-syn}
\left\{
\begin{aligned}
X_1 & \sim  Pois(65+0.5\cdot E) \\
X_2 & \sim Bernoulli(0.3-0.025\cdot E) \\
X_3 & \sim Bernoulli(0.2) \\
X_4 & \sim Bernoulli(\sigma(-0.5+0.2\cdot E+1.3\cdot X_3)) \\
X_5 & \sim Bernoulli(\sigma(-1+0.3\cdot E+0.015\cdot X_1+0.001\cdot X_2 +1.5\cdot X_3)) \\
X_6 & \sim Bernoulli(0.175-0.015\cdot E) \\
X_7 & \sim Bernoulli(0.3) \cdot \mathbf{I}_{X_6=0}  \\
X_8 & \sim Bernoulli(0.6) \cdot \mathbf{I}_{X_6+X_7=0} \\
T & \sim log\mathcal{N} (1.5+0.4E-0.1(X_1-65)-0.05X_2-1.75X_3-2.5X_4\\
&+0.6X_5+0.25X_6-0.75X_7-2X_8,1) \\
Y & =  \mathbf{I}_{T > 5}
\end{aligned} \right.
\end{equation}
where $T$ is an intermediate variable for deriving $Y$, which will not show as a feature in the dataset, $\log \mathcal{N}(\cdot,\cdot)$ stand for the log-normal distribution. 

\subsubsection{Regression} 
\begin{figure}[h!]
	\centering
	\includegraphics[scale=0.5]{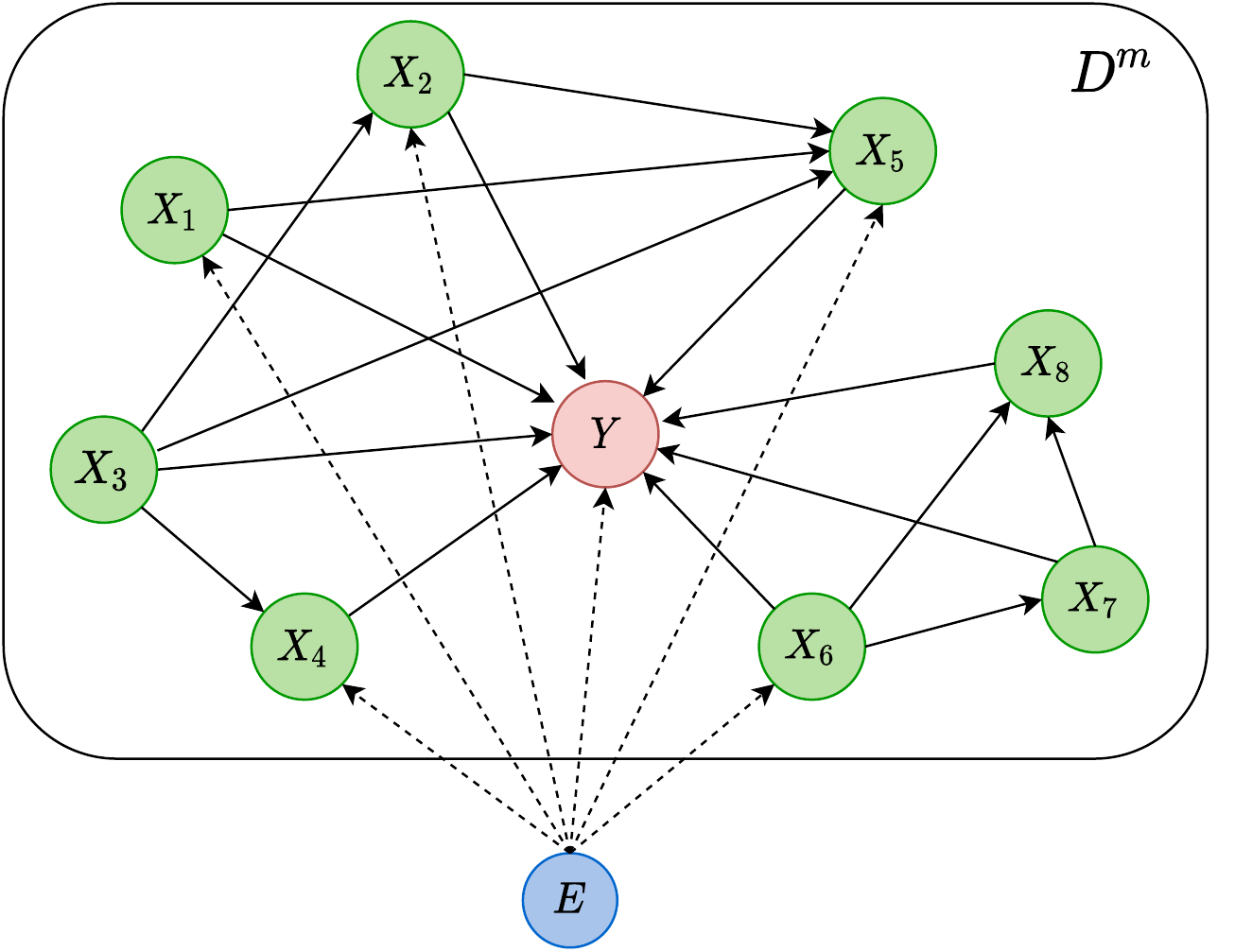}
	\includegraphics[scale=0.55]{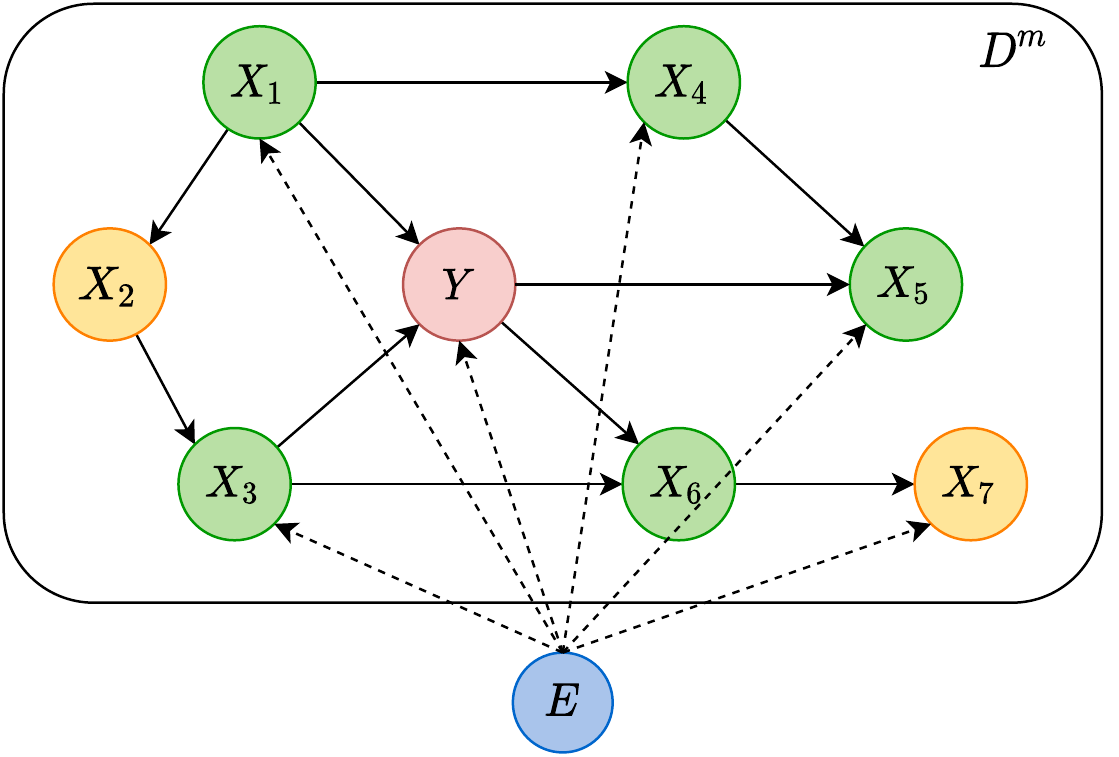}
	\caption[The DAG for synthetic datasets]{The underlying DAG for the synthetic classification (left) and regression (right) datasets}
	\label{syndag}
\end{figure}

The second dataset for regression task is generated according to the DAG in Figure \ref{syndag}. Its structural equations are sketched below: 
\begin{equation}
\label{reg-syn}
\left\{
\begin{aligned}
X_1 & = 0.8 E + \epsilon_1\\
X_2 & = 0.4 X_1^2 + \epsilon_2 \\
X_3 &= 0.3 E + 0.1 \exp(X_2) + \epsilon_3 \\
Y &= -0.5 E^2 + \log(0.3X_1^2+0.7X_2^2) + \epsilon_y \\
X_4 &= 0.1X_1\cdot \sqrt{\exp(E)} + \epsilon_4 \\
X_5 &= -0.25E\cdot X_4 + 0.6Y + \epsilon_5 \\
X_6 &= -1 + 0.2X_3 \cdot Y  + \epsilon_6 \\
X_7 &= -0.6E + 3X_6 + \epsilon_7
\end{aligned} \right.
\end{equation}

\subsection{Verifying Intuition on Synthetic Datasets}

In this section, we verify the close relationship between $E$ difference and Wasserstein distance of two distributions through synthetic causal data and meanwhile dive deeper into how well DAPDAG can learn this $E$ and exploit this for domain adaptation. 

\paragraph{Wasserstein Distance between two empirical distributions} 

\begin{figure}[h!]
	\centering
	\includegraphics[scale=0.9]{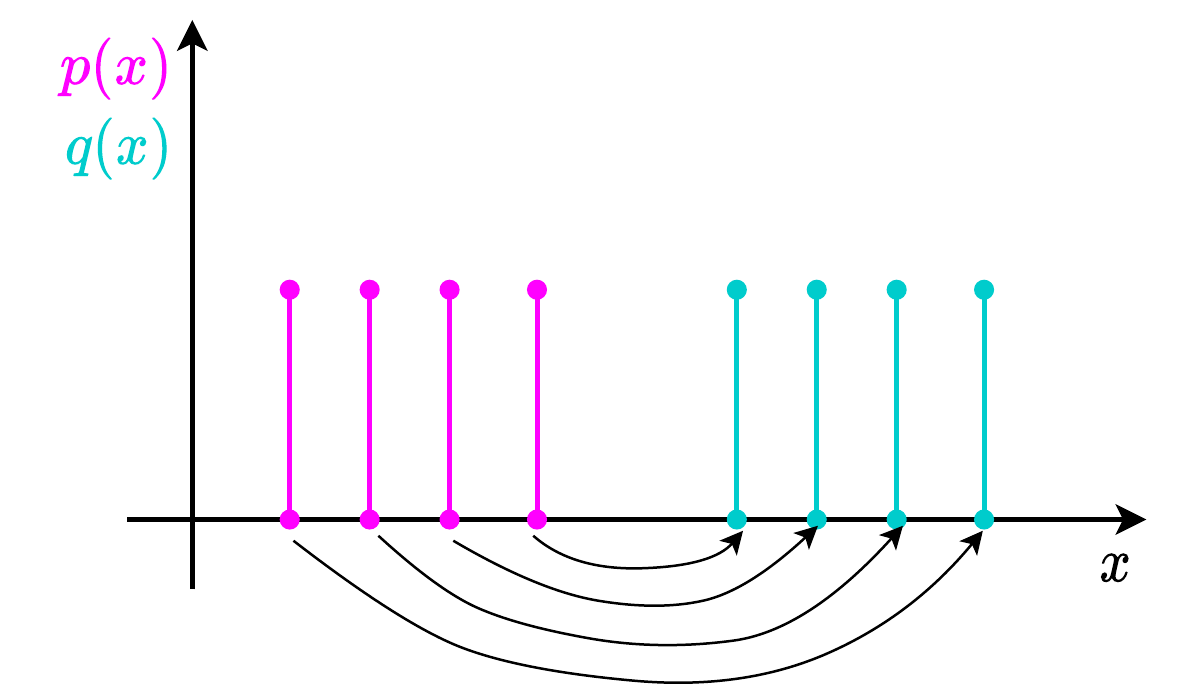}
	\label{wasser}
	\caption[Simple Example of Transport between Two Empirical Distributions]{Simple Example of Transport between Two Empirical Distributions}
\end{figure}

There exist extensive works inspecting distribution distances, e.g. KL-divergence and H-divergence, and how to utilise these metrics for further applications. In our work, we use a distance metric called Wasserstein distance to measure the distance of two empirical distributions \citep{panaretos2019statistical}. It's formal mathematical definition is below:
The $p$-Wasserstein distance between probability measures $\mu$ and $\nu$ on $\mathbb{R}^d$ is defined as
\begin{equation}
W_p(\mu,\nu) = \inf\limits_{X\sim\mu,Y\sim\nu}(\mathbb{E}||X-Y||^p)^{\frac{1}{p}}, \quad p \geq 1.
\end{equation}

A very high-level understanding of the distance metric from the optimal transport perspective is the minimum effort it would take to move points of mass from one distribution to the other. Let's consider a simple example in Figure \ref{wasser} where we want to move the points in $p(x)$ to the same places of points in $q(x)$. There can be a lot of ways of moving, and the arrows in the Figure depict one of them. However, what we are interested is the way with the least effort. This can be approximated using a numeric method called Sinkhorn iterations \citep{cuturi2013sinkhorn}. Since our focus is on DA, we skip the details of this algorithm and directly apply the method to compute the distance of each pair of synthetic datasets.           

\paragraph{Visualisation of Results}
\begin{figure}[h!]
	\centering
	\includegraphics[scale=0.95]{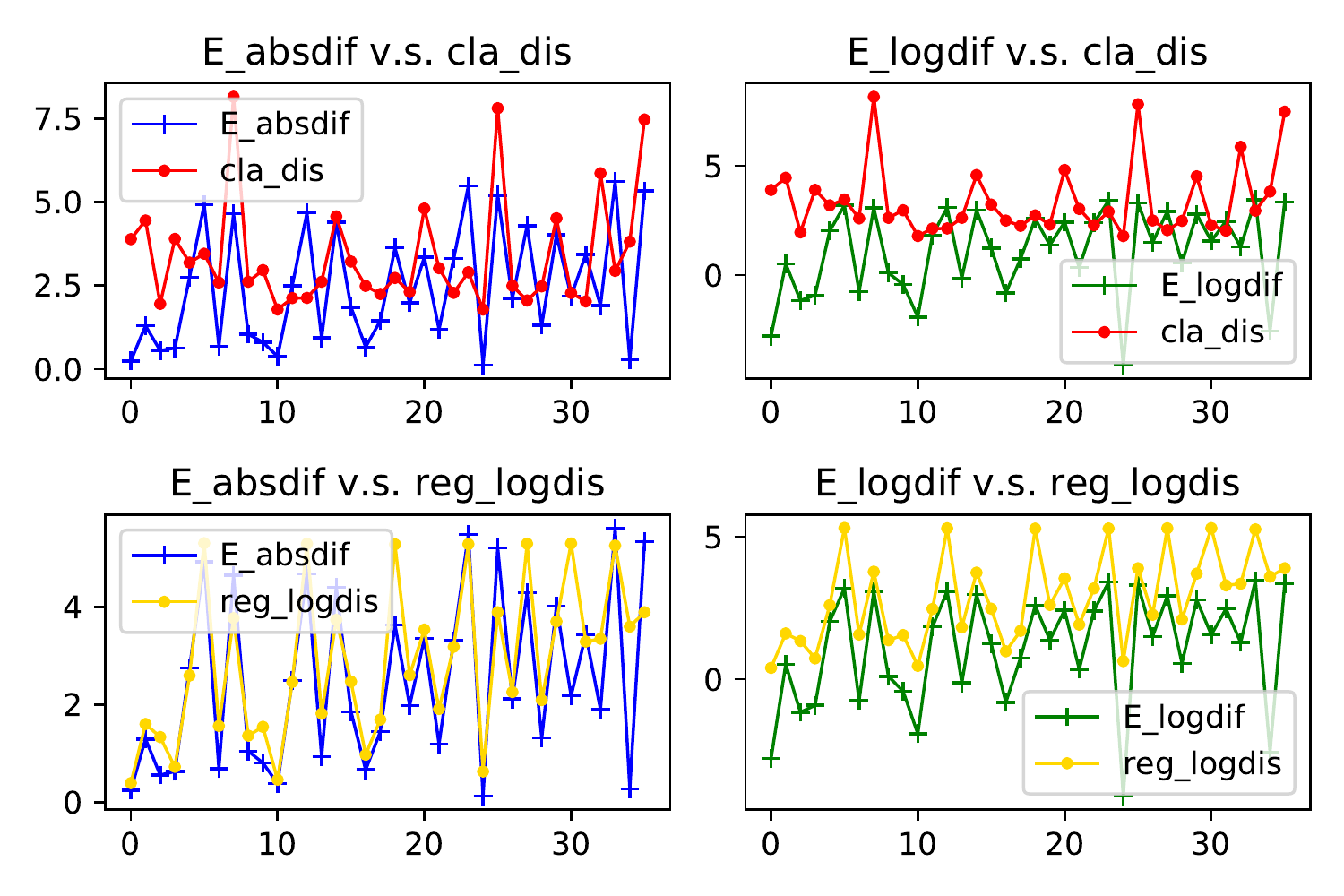}
	\label{e-dis}
	\caption[Comparisons between E differences and Sinkhorn distances of synthetic classification and regression datasets]{Comparisons between E differences and Sinkhorn distances of synthetic classification and regression datasets (numbers in the x-axis stand for the indices of domain pairs, e.g. if we have total 9 domains, we will have 36 domain pairs indexed from 0 to 35.)}
\end{figure}
\begin{figure}[h!]
	\centering
	\includegraphics[scale=0.9]{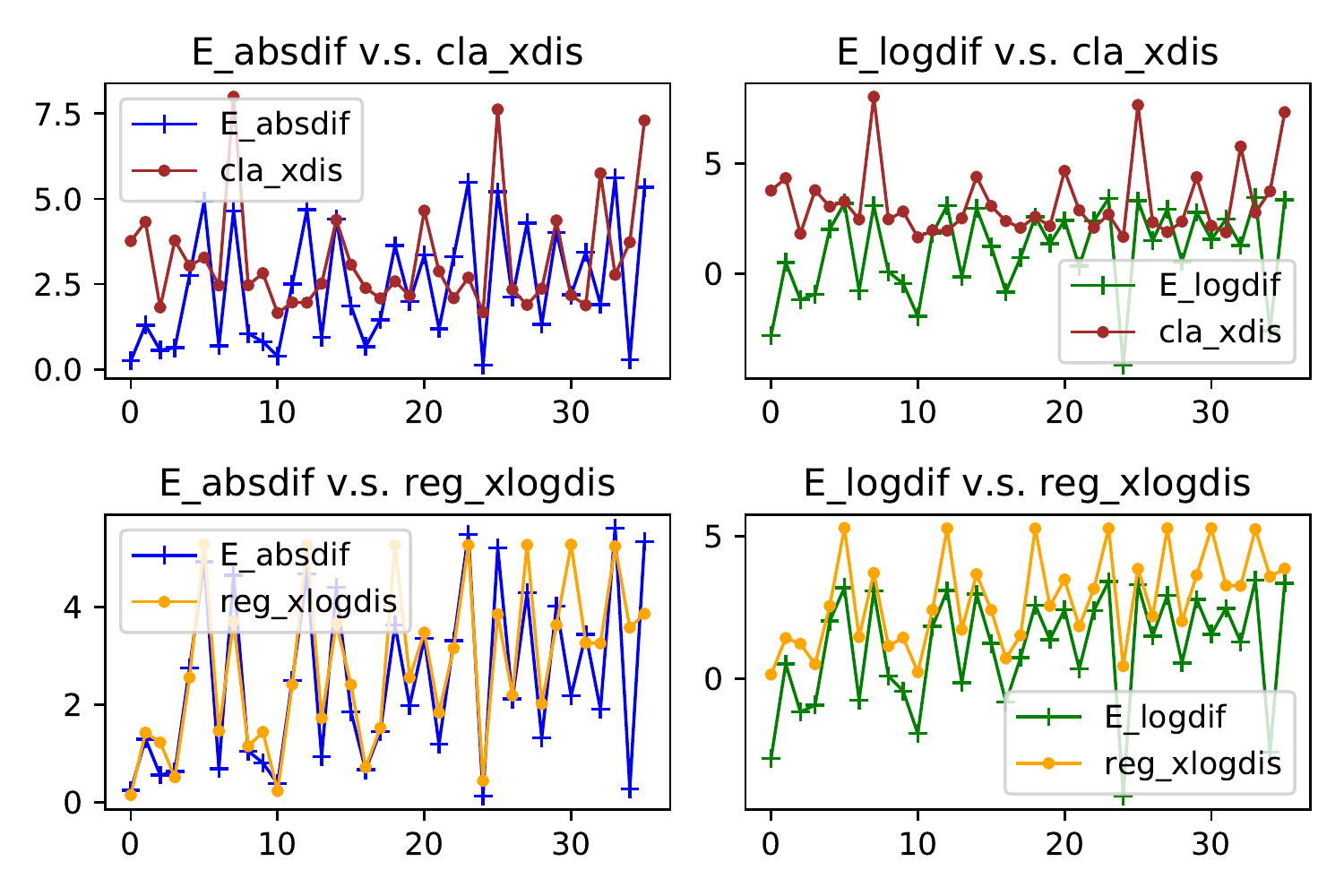}
	\label{e-xdis}
	\caption[Comparisons between E differences and Sinkhorn distances of features from synthetic classification and regression datasets]{Comparisons between E differences and Sinkhorn distances of features from synthetic classification and regression datasets}
\end{figure}

It is an interesting fact from Figure \ref{e-dis} that the difference of $E$s that are used for generating two synthetic datasets can be a regarded as good proxy of Wasserstein distance between these two datasets. For regression data, the absolute difference of $E$s almost fully coincides with the log of Wasserstein distance in terms of both values and fluctuations. Since our method utilises the features in the target domain for adaptation, we also plot the relationship between $E$ difference and Wasserstein distance of features in \ref{e-xdis}. As shown in the plots, ignorance of labels barely affects the relationship. This finding provides a strong evidence for using only features to find the distribution difference and adjust for the shift accordingly.   

\begin{figure}[h!]
	\centering
	\includegraphics[scale=0.9]{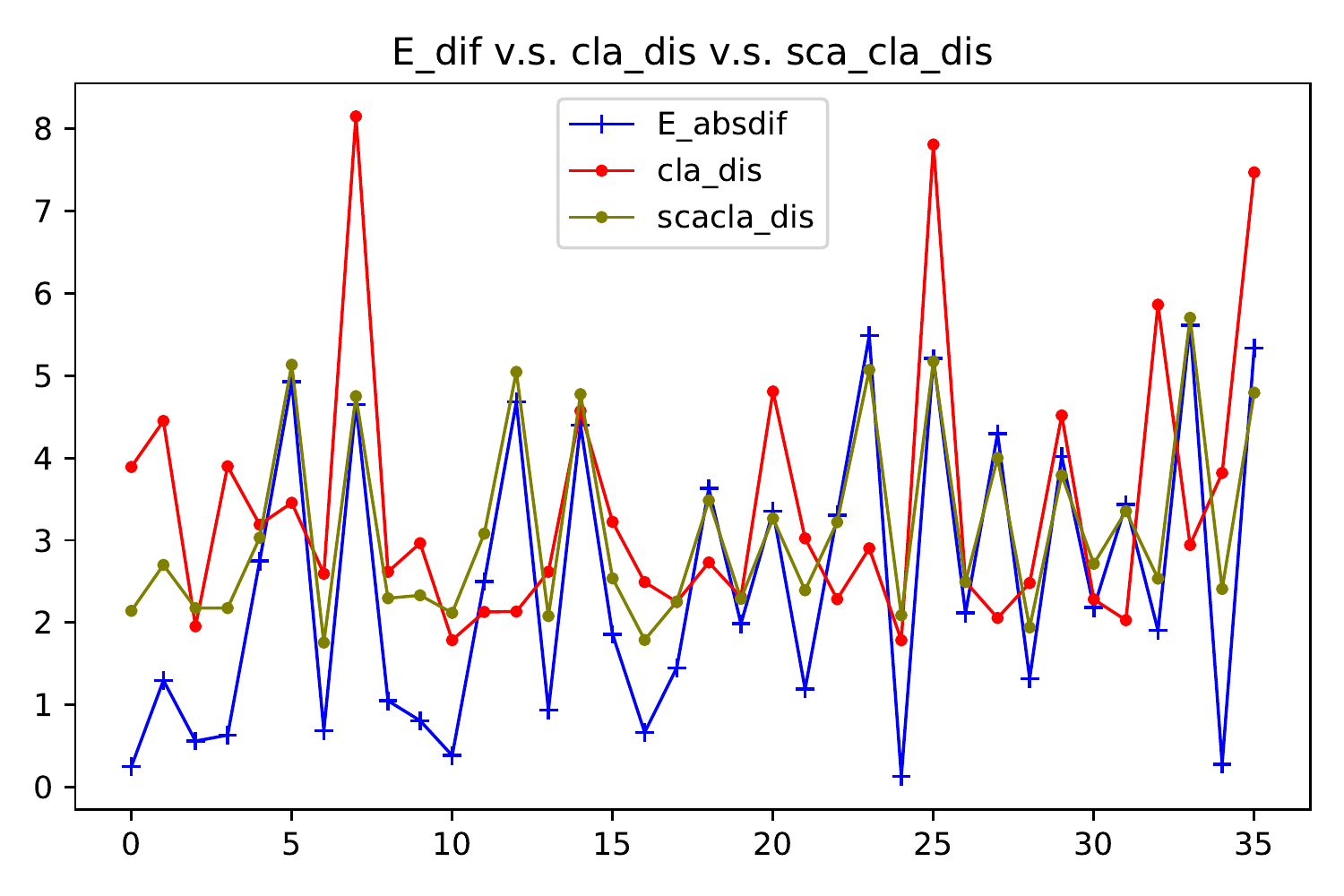}
	\label{e-scadis}
	\caption[Comparisons between E differences and Sinkhorn distances of synthetic classification datasets with and without standardisation]{Comparisons between E differences and Sinkhorn distances of synthetic classification datasets with and without standardisation.}
\end{figure}

However, on the classification dataset, we can see that despite of the resemblance on fluctuations, true values of $E$ differences deviate to some extent from distances of both full variables and only features. Luckily, we can relieve this issue by standardising the features with large scales. And after standardisation, the distance can better capture the variation of $E$ difference, as illustrated in Figure \ref{e-scadis}.

\begin{figure}[h!]
	\centering
	\includegraphics[scale=0.55]{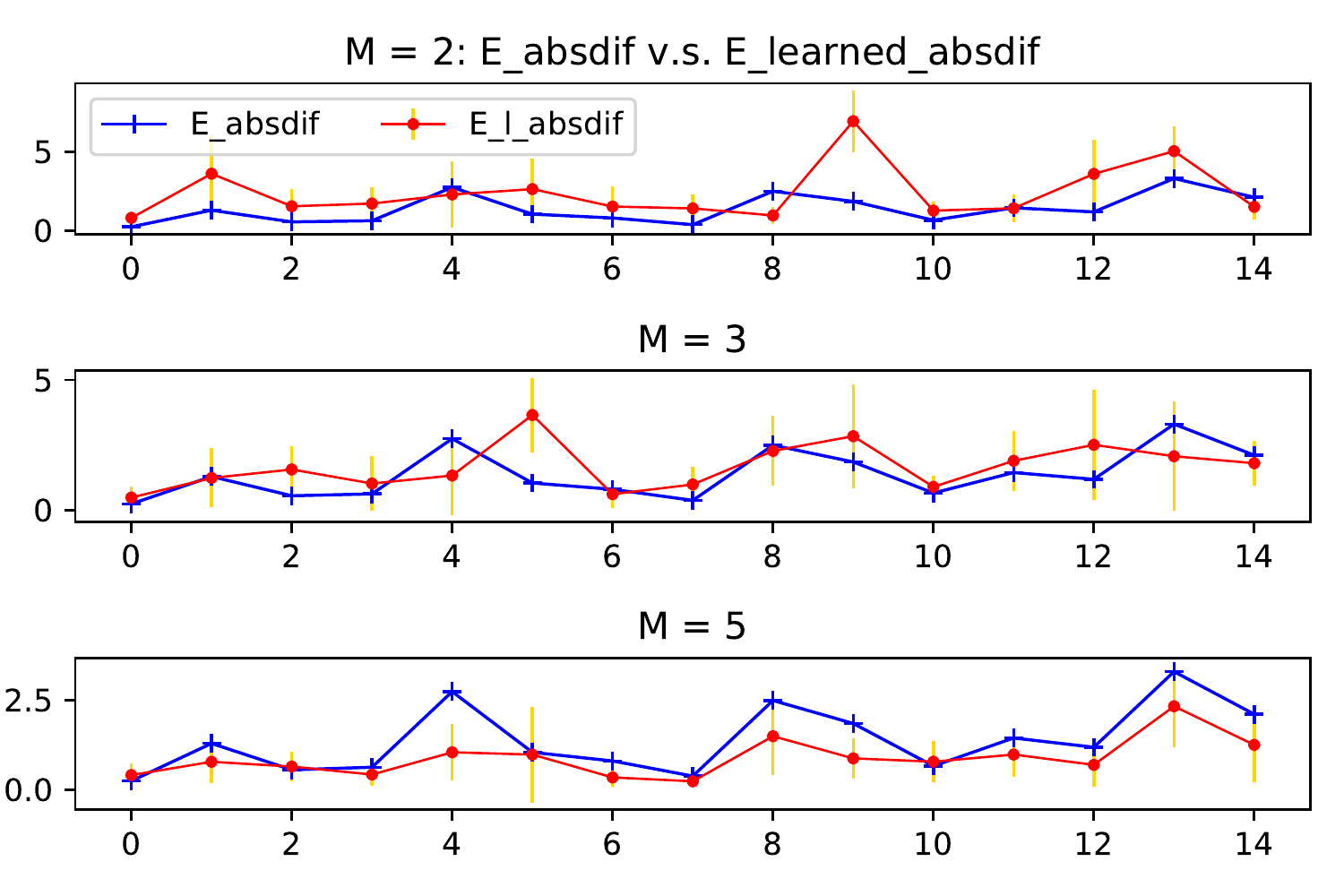}
	\label{e-learned}
	\caption[Comparisons between E differences and learned E differences from synthetic regression datasets with different number of source domains used]{Comparisons between E differences and learned E differences from synthetic regression datasets with different number of source domains used (M is the number of source domains and numbers in the x-axis stand for the indices of domain pairs, here for better graphical presentation, we show 15 randomly-selected domain pairs).}
\end{figure}

\paragraph{Capturing the $E$ difference}
How well can our method learn the $E$ difference so as to enable its ability of domain adaptation? We observe that as the number of sources increases, the learned $E$ difference catches better the trend of true $E$ difference, which is exhibited in Figure \ref{e-learned}. This supports the benefit of training more sources for adaptation.

\subsection{Ablation Studies}
\label{abl}
We do ablation studies on various loss components in (\ref{losscomp}) except for the regularisation loss on $E$ to better understand sources of performance gain. It is noticed that the comparison experiment with CASTLE can be considered as an ablation study on the encoder and once this $E$ is introduced, the square of $E$ should be regularised as proved in \ref{pac-bound} to lower the generalisation bound of decoder during training. Therefore, it is not necessary to do a separate ablation study on the squared regularisation term in \ref{loss}. Besides, we have shown the comparison with BRM as an ablation study on structural filters. Both DAPDAG and CASTLE have the same number of structural filters as the total number of variables and these filters contribute to the reconstruction of each variable and DAG learning. In BRM, however, there is only one filter that selects features locally for the target variable. 

\begin{table}[h!]
	\centering
	\begin{tabular}{ |c|c|c|c|c|c|c|}
		\hline
		\multirow{2}{3em}{Methods} &\multicolumn{2}{|c|}{M=3} & \multicolumn{2}{|c|}{M=5} &\multicolumn{2}{|c|}{M=7}  \\
		\cline{2-7}
		& AUC & APR & AUR & APR & AUR & APR  \\
		\hline
		Dag+Spa & 0.947(0.021) & 0.814(0.091) & 0.954(0.017)& 0.820(0.075) & 0.958(0.015) & 0.827(0.063)   \\
		Rec+Dag & 0.959(0.006) & 0.825(0.072) & 0.961(0.004) &  0.849(0.063)  & 0.962(0.004) & 0.872(0.049)  \\
		Rec+Spa & 0.960(0.005) & 0.845(0.069) & 0.962(0.005)&  0.854(0.055)& 0.963(0.003) & 0.890(0.044)  \\
		Rec+Dag+Spa & \textbf{0.961(0.004)}& \textbf{0.856(0.036)} & \textbf{0.964(0.004)}& \textbf{0.883(0.033)} & \textbf{0.965(0.003)} & \textbf{0.893(0.031)}  \\
		\hline
	\end{tabular}
	\caption[Ablation studies of DAPDAG on synthetic classification dataset]{Ablation studies on synthetic classification dataset. M is the number of source domains and evaluation metric scores of AUR and APR are averaged over multiple runs with the respective standard deviation in the parentheses, in each of which a target and source domains are randomly selected from a pool of domains. (\textbf{Dag}: NOTEARS regularisation; \textbf{Spa}: group-lasso regularisation on the structural filters; \textbf{Rec}: reconstruction loss of all observed variables.)}
	\label{syn-cla-abl}
\end{table}

\begin{table}[h!]
	\centering
	\begin{tabular}{ |c|c|c|c|  }
		\hline
		\multirow{2}{3em}{Methods} &\multicolumn{3}{|c|}{RMSE}  \\
		\cline{2-4} 
		& M=3 & M=5 & M=7 \\
		\hline
		Dag+Spa & 0.422(0.325) & 0.444(0.306) & 0.495(0.258) \\
		Rec+Dag & 0.479(0.254) & 0.508(0.221) & 0.545(0.173) \\
		Rec+Spa & 0.486(0.231) & 0.510(0.209) & 0.558(0.166) \\
		Rec+Dag+Spa & \textbf{0.501(0.200)} & \textbf{0.533(0.167)} &\textbf{0.572(0.142)} \\
		\hline
	\end{tabular}
	\caption[Ablation studies of DAPDAG on synthetic regression dataset]{Ablation studies on synthetic regression dataset. M is the number of source domains and the $R^2$ scores are averaged over multiple runs with the respective standard deviation in the parentheses, in each of which a target and source domains are randomly selected from a pool of domains. (Dag: NOTEARS regularisation; Spa: group-lasso regularisation on the structural filters; Rec: reconstruction loss of all observed variables.)}
	\label{syn-reg-abl}
\end{table}

\begin{figure}[h!]
	\centering
	\includegraphics[scale=0.55]{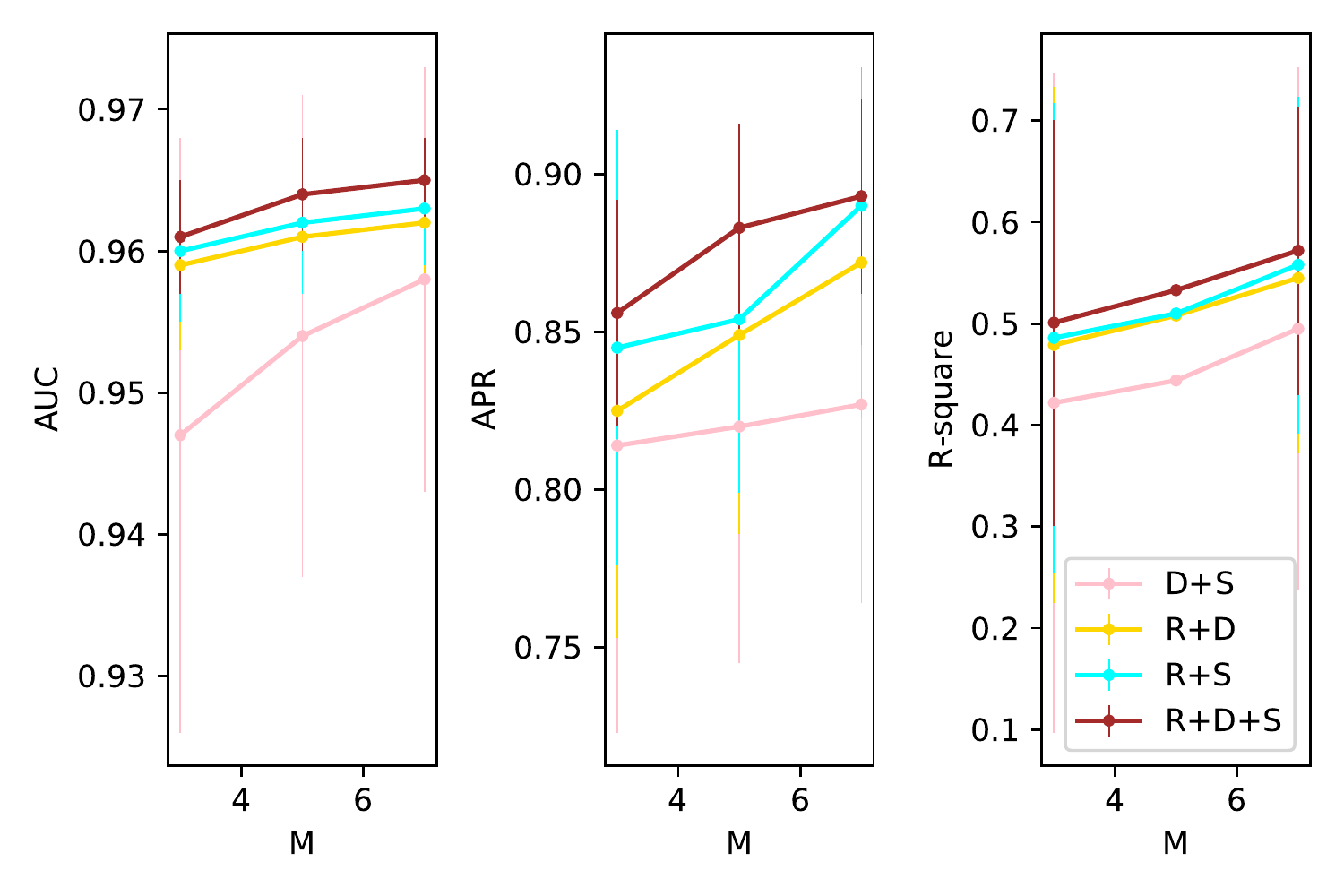}
	\caption{Ablation Studies on Synthetic Datasets. In ablation studies, R stands for only including reconstruction loss for the DAG loss; D stands for only including DAG constraint for the DAG loss; S stands for only including sparsity loss for the DAG loss;}
	\label{fig:synabl}
\end{figure}

The comparison results in Figure \ref{fig:synabl} verify the importance of structural filters and the regularisation on these filters as a DAG. On the regression dataset, if we do not reconstruct each variable, the performance of DAPDAG will be even worse than BRM with much simpler structure. Therefore, reconstruction of all variables brings significant benefit to prediction while DAG and sparsity constraint further improves the model's robustness across different domains. 

\subsection{Scalability}
\label{sca}
\begin{figure}[h!]
	\centering
	\includegraphics[scale=0.6]{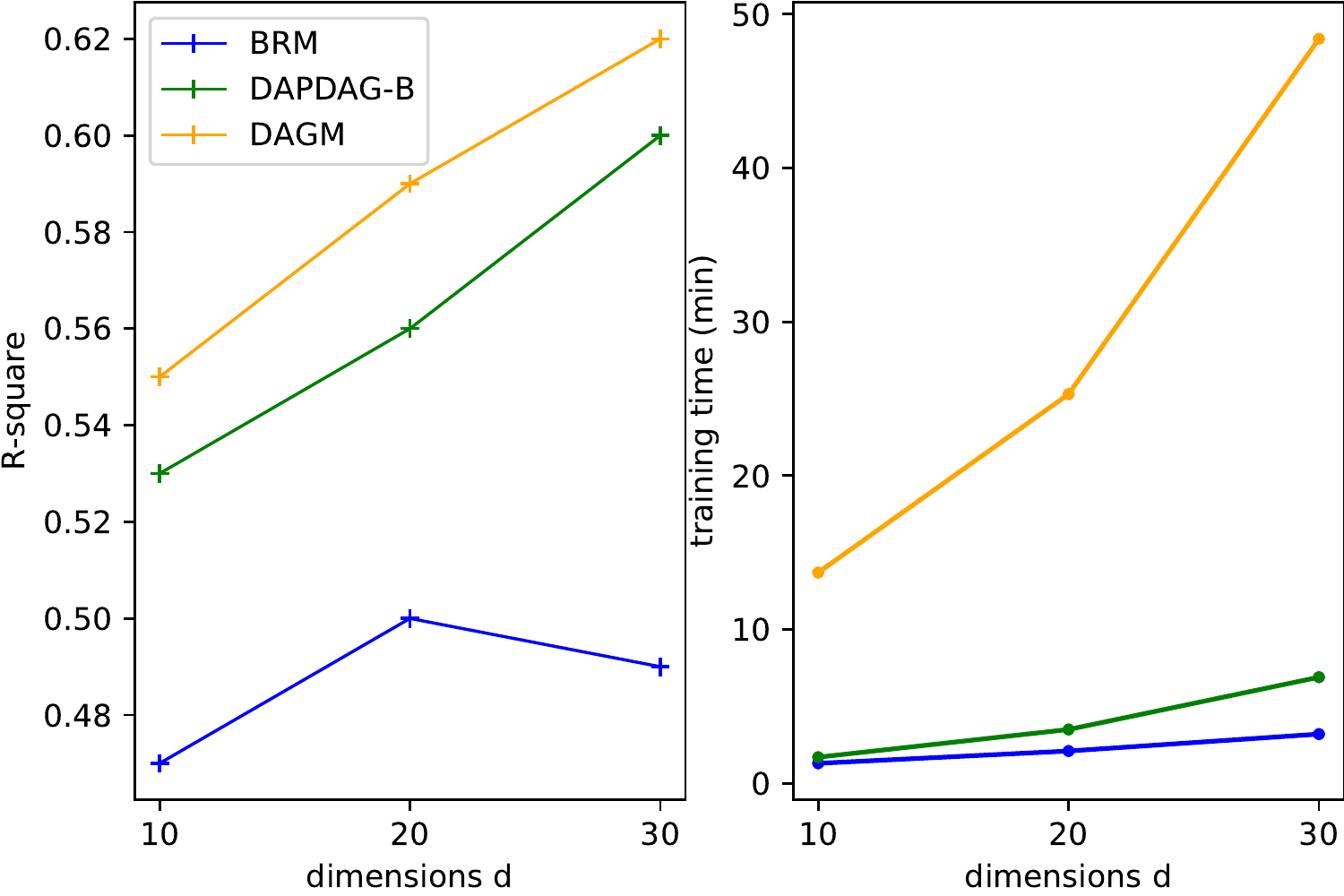}
	\caption{Performance and Scalability}
	\label{fig:sca}
\end{figure}

We have extended simulations to cases with higher dimensions, about which you may find more information in Figure \ref{fig:sca}. In the right plot, we compare our method with two UDA benchmarks in training time versus data dimensions. Despite the minor gap between our approach and \cite{zhang2020domain}, ours needs considerately less time for training than theirs in high-dimensional settings. 

\subsection{Processing of MAGGIC Dataset}
It is tricky that data-preprocessing and domain selection can exert considerate influence on testing performance because these datasets have extensive missing values or features in each study while the usual data imputation methods tend to impute those missing values without taking account of domain distinction. And we admit that in this part future work is needed for better data imputation or UDA methodology that can deal with feature mismatch.

\paragraph{Imputation of Missing Values} 
Despite extensive instances contained by MAGGIC, each study tends to have massive missing values in certain features or even a number of missing features, which significantly violates our assumptions in terms of causal sufficiency and feature match. Hence it is imperative to process these datasets before use. We omit those with missing labels and use MissForest \citep{stekhoven2012missforest}, a non-parametric missing value imputation method for mixed-type data, to impute missing values of features. We first iterate imputation of missing values over studies. During imputing missing entries in each study, we try to rely on other features of that study as much as possible. For missing features that cannot be imputed based on the single study, we resort to other studies that have the features available. For binary features, the imputed values would be fractions between 0 and 1, which are transformed to 0 or 1 with a threshold at 0.5. 

\paragraph{Selection of Domains}
We then select processed studies with fewest missing features originally for experiments because those studies are supposed to be affected least by data imputation and maintain the distribution shift from other domains. The selected studies are shown in Table \ref{maggic-M10-prop} with dataset size followed in the parentheses.

\paragraph{Standardisation of Non-binary features}
All continuous features are standardised with mean of 0 and variance of 1, just following the same procedure as we do for synthetic classification dataset.

Meanwhile, a recent work \citep{kaiser2021unsuitability} claims that continuous optimisation/differentiable methods of causal discovery such as NO-TEARS may not work well on dataset with variant scales. They observe inconsistent learning results with respect to data scaling - variables with larger scales or variance tend to be the child nodes. Standardising the data with large scales can alleviate the problem to some extent.

\subsection{Supporting Experimental Figures and Plots}
\label{supp}

\subsubsection{Comparison against Benchmarks on MAGGIC Datasets}
\begin{table}[h!]
	\centering
	\begin{tabular}{ |c|c|c|c|c|c|c|  }
		\hline
		\multirow{2}{3em}{Target Study} & \multicolumn{6}{|c|}{AUC Scores} \\
		\cline{2-7}
		& Deep MLP  & MDAN  & BRM & MisForest & CASTLE & DAPDAG-B  \\
		\hline
		\textbf{AHFMS} (99.7) & 0.782(0.012) & 0.785(0.013) & 0.811(0.010) & 0.819(0.006) & 0.826(0.019) & \textbf{0.854(0.011)}  \\
		\textbf{BATTL} (58.8) & 0.692(0.016) & 0.707(0.020) & 0.735(0.014)& 0.765(0.007) & 0.768(0.013) & \textbf{0.790(0.012)} \\
		\textbf{Hilli} (111.5) & 0.687(0.014) & 0.695(0.020) & 0.699(0.014) & 0.711(0.012) & 0.713(0.005) & \textbf{0.730(0.013)} \\
		\textbf{Kirk} (46.9) & 0.868(0.005) & 0.891(0.009) & 0.931(0.012) & 0.936(0.010) & 0.954(0.008) & \textbf{0.970(0.009)} \\
		\textbf{Macin} (361) & 0.569(0.010) & 0.567(0.007) & 0.619(0.015) & 0.588(0.014) & 0.625(0.017) &  \textbf{0.646(0.015)} \\
		\textbf{Mim B} (59.4) & 0.596(0.011) & 0.578(0.014) & 0.612(0.018) & 0.647(0.013) & 0.616(0.024) &  \textbf{0.659(0.016)} \\
		\textbf{NPC I} (71.5) & 0.517(0.016) & 0.524(0.011) & 0.540(0.017) & 0.542(0.021) & 0.533(0.011) & \textbf{0.571(0.020)} \\
		\textbf{Richa} (36.6) & 0.712(0.012) & 0.711(0.013) & 0.739(0.013) & \textbf{0.782(0.011)} & 0.757(0.012) &  0.775(0.017) \\
		\textbf{SCR A} (54.9) & 0.706(0.017) & 0.698(0.024) & 0.710(0.019) & 0.675(0.009) & 0.691(0.022) &  \textbf{0.728(0.018)} \\
		\textbf{Tribo} (56.6) & 0.760(0.006) & 0.771(0.010) & 0.766(0.016) & 0.769(0.012) & 0.788(0.015) & \textbf{0.799(0.010)} \\
		\hline 	
	\end{tabular}
	\caption[Classification AUC of DAPDAG-B on MAGGIC Dataset against other benchmarks for each target study]{Classification performance of DAPDAG-B on MAGGIC Dataset against other benchmarks for each target study in the selection pool. For each target study, we set corresponding training domains to be rest 9 studies in the selection pool and add its average distance with respect to sources behind its name in the first column. The performance scores are the averaged AUC over 10 replicates, with standard deviation in the parentheses. Bold denotes the best.}
	\label{maggic-M10-prop}
\end{table}

\begin{table}[h!]
	\centering
	\begin{tabular}{ |c|c|c|c|c|c|c| }
		\hline
		\multirow{2}{3em}{Target Study} & \multicolumn{6}{|c|}{APR Scores} \\
		\cline{2-7}
		& Deep MLP  & MDAN  & BRM & MisForest & CASTLE & DAPDAG  \\
		\hline
		\textbf{AHFMS} (196) & 0.914(0.011) & 0.921(0.008)  & 0.930(0.014) & 0.936(0.007) & 0.938(0.009) & \textbf{0.949(0.009)}  \\
		\textbf{BATTL} (363)  & 0.947(0.013) & 0.953(0.010) & 0.955(0.009) & 0.965(0.003) & 0.966(0.004)  & \textbf{0.970(0.006)} \\
		\textbf{Hilli} (176) & 0.853(0.007) & 0.865(0.013) & 0.866(0.010) & 0.869(0.006) & 0.881(0.002) &  \textbf{0.897(0.008)}\\
		\textbf{Kirk} (215)  & 0.923(0.007) & 0.938(0.006) & 0.952(0.012) & 0.967(0.004) & 0.969(0.002) &  \textbf{0.972(0.007)} \\
		\textbf{Macin} (228) & 0.506(0.012) & 0.514(0.019) & 0.517(0.017) & 0.497(0.017) & 0.547(0.014) & \textbf{0.581(0.016)} \\
		\textbf{Mim B} (282) & 0.812(0.009) & 0.821(0.014) & 0.825(0.016) & 0.837(0.007) & 0.819(0.021) & \textbf{0.846(0.013)} \\
		\textbf{NPC I} (66) & 0.528(0.011) & 0.529(0.019) & 0.551(0.018) & 0.546(0.013) & 0.565(0.023) & \textbf{0.569(0.018)} \\
		\textbf{Richa} (627) & 0.879(0.008) & 0.884(0.011) & 0.894(0.011) & \textbf{0.921(0.010)} & 0.912(0.006) & 0.918(0.010) \\
		\textbf{SCR A} (324) & 0.959(0.011) & 0.952(0.012) & 0.963(0.013) & 0.965(0.008) & 0.967(0.013) & \textbf{0.975(0.012)} \\
		\textbf{Tribo} (663) & 0.914(0.005) & 0.920(0.014) & 0.927(0.012) & 0.921(0.009) & 0.935(0.010) & \textbf{0.939(0.011)} \\
		\hline 	
	\end{tabular}
	\caption[Classification APR of DAPDAG on MAGGIC Dataset against other benchmarks for each target study]{Classification performance of DAPDAG-B on MAGGIC Dataset against other benchmarks for each target study in the selection pool. For each target study, we set corresponding training domains to be rest 9 studies in the selection pool and add its sample size behind its name in the first column. The performance scores are the averaged APR over 10 replicates, with standard deviation in the parentheses. Bold denotes the best.}
	\label{maggic-M10-prop2}
\end{table}

\vfill

\end{document}